\magnification=\magstep0
\hsize=6.5 true in
\vsize=9.0 true in
\overfullrule=0pt
\baselineskip=15pt
\tolerance=10000
\hbadness=200

\font\tensc=cmcsc10
\newfam\scfam
\textfont\scfam=\tensc
\def\sc{\fam\scfam\tensc}
 
\font\bigbf=cmbx10 scaled\magstep1

\font\biggbf=cmbx10 scaled\magstep2

\font\bigggbf=cmbx10 scaled\magstep3

\font\bigit=cmti10 scaled\magstep1

\def\Kappa{{\rm K}}

\def\sectionbreak{
\bigskip\vskip\parskip}

\def\bigsectionbreak{
\bigskip\bigskip\vskip\parskip}

\outer\def\beginsection#1\par
{\sectionbreak
\message{#1}\leftline{\bf#1}\nobreak\smallskip\noindent}

\outer\def\bigbeginsection#1\par
{\sectionbreak
\message{#1}\leftline{\bigbf#1}\nobreak\medskip\noindent}

\def\currentsection{\firstmark}

\outer\def\biggbeginsection#1\par
{\bigsectionbreak
\message{#1}\leftline{\biggbf#1}
\mark{#1}\nobreak\bigskip\noindent}

\outer\def\bigitbeginsection#1\par
{\sectionbreak
\message{#1}\leftline{\bigit#1}\nobreak\medskip\noindent}

\outer\def\longbigbeginsection#1 #2\par#3\par
{\sectionbreak
\message{#1 #2 #3}
\halign{##\hfil&##\hfil\cr
{\bigbf#1\ }&{\bigbf#2}\cr
&{\bigbf#3}\cr
}\nobreak\medskip\noindent}

\outer\def\longbiggbeginsection#1 #2\par#3\par
{\bigsectionbreak
\message{#1 #2 #3}
\halign{##\hfil&##\hfil\cr
{\biggbf#1\ }&{\biggbf#2}\cr
&{\biggbf#3}\cr
}\mark{#1 #2 #3}\nobreak\bigskip\noindent}

\outer\def\longbiggbeginappendix#1 #2 #3\par#4\par
{\bigsectionbreak
\message{#1 #2 #3 #4}
\halign{##\hfil&##\hfil\cr
{\biggbf#1\ #2\ }&{\biggbf#3}\cr
&{\biggbf#4}\cr
}\mark{#1 #2 #3 #4}\nobreak\bigskip\noindent}

\def\rightheadline{\hfil{\it\currentsection}\ \ \ \ \ \ {\rm \folio}}

\def\currentchapter{}

\def\leftheadline{{\rm \folio}\ \ \ \ \ \ {\it\currentchapter}\hfil}

\newcount\titlepageno

\def\setheadline{\headline=
{\ifnum\titlepageno=\pageno{\hfil}
\else{\ifodd\pageno{\rightheadline}\else{\leftheadline}\fi}
\fi}}

\def\skipifeven
{\ifodd\pageno{}\else\advancepageno\fi}

\outer\def\beginchapter#1. #2\par
{\vfill\eject\skipifeven\titlepageno=\pageno
\def\currentchapter{Cap\'\i tulo #1. #2}
\topinsert\vskip 0.25\vsize\endinsert
\hrule\medskip
\rightline{\bigggbf Cap\'\i tulo #1}
\medskip
\rightline{\bigggbf#2}
\medskip\hrule\bigskip\bigskip}

\outer\def\shortbeginchapter#1\par
{\vfill\eject\skipifeven\titlepageno=\pageno
\def\currentchapter{#1}
\mark{#1}
\topinsert\vskip 0.25\vsize\endinsert
\hrule\medskip
\rightline{\bigggbf#1}
\medskip\hrule\bigskip\bigskip}

\def\itemitemitem{\par\indent\indent\hangindent3\parindent
\textindent}

\def\itemitemitemitem{\par\indent\indent\indent\hangindent4
\parindent\textindent}

\def\LaTeX{{\rm L\kern-.36em\raise.3ex\hbox{\sc a}\kern-.15em%
    T\kern-.1667em\lower.7ex\hbox{E}\kern-.125emX}}

\def\mytilde{\kern-.5pt\lower3pt\hbox{\char'176}\kern.5pt}

\input epsf


\catcode`@=11 \catcode`!=11

\expandafter\ifx\csname fiverm\endcsname\relax
  \let\fiverm\fivrm
\fi
  
\let\!latexendpicture=\endpicture 
\let\!latexframe=\frame
\let\!latexlinethickness=\linethickness
\let\!latexmultiput=\multiput
\let\!latexput=\put
 
\def\@picture(#1,#2)(#3,#4){%
  \@picht #2\unitlength
  \setbox\@picbox\hbox to #1\unitlength\bgroup 
  \let\endpicture=\!latexendpicture
  \let\frame=\!latexframe
  \let\linethickness=\!latexlinethickness
  \let\multiput=\!latexmultiput
  \let\put=\!latexput
  \hskip -#3\unitlength \lower #4\unitlength \hbox\bgroup}

\catcode`@=12 \catcode`!=12

\input pictex

 
\catcode`@=11 \catcode`!=11
  
\let\!pictexendpicture=\endpicture 
\let\!pictexframe=\frame
\let\!pictexlinethickness=\linethickness
\let\!pictexmultiput=\multiput
\let\!pictexput=\put

\def\beginpicture{%
  \setbox\!picbox=\hbox\bgroup%
  \let\endpicture=\!pictexendpicture
  \let\frame=\!pictexframe
  \let\linethickness=\!pictexlinethickness
  \let\multiput=\!pictexmultiput
  \let\put=\!pictexput
  \let\input=\@@input   
  \!xleft=\maxdimen  
  \!xright=-\maxdimen
  \!ybot=\maxdimen
  \!ytop=-\maxdimen}

\let\frame=\!latexframe

\let\pictexframe=\!pictexframe

\let\linethickness=\!latexlinethickness
\let\pictexlinethickness=\!pictexlinethickness

\let\\=\@normalcr
\catcode`@=12 \catcode`!=12

\newdimen\beforeamount\beforeamount=10pt
\newdimen\betweenamount\betweenamount=15pt
\newdimen\afteramount\afteramount=10pt
\newdimen\digitwidth\digitwidth=5pt
\def\skipbefore{\hskip\beforeamount}
\def\skipbetween{\hskip\betweenamount}
\def\skipafter{\hskip\afteramount}
\def\skipd{\hskip\digitwidth}

{\baselineskip=21.6pt
\centerline{\biggbf Two Novel Evolutionary  Formulations}
\centerline{\biggbf of the Graph Coloring Problem}
}

\bigskip\bigskip
\centerline{{\it Valmir C. Barbosa\/}$^*$}
\centerline{{\it Carlos A. G. Assis\/}$^*$}
\centerline{{\it Josina O. do Nascimento\/}$^{*\dag}$}

\bigskip
\centerline{$^*$\enspace Universidade Federal do Rio de Janeiro}
\centerline{Programa de Engenharia de Sistemas e Computa\c c\~ao, COPPE}
\centerline{Caixa Postal 68511, 21945-970 Rio de Janeiro - RJ, Brazil}

\bigskip
\centerline{$^{\dag}$\enspace Observat\'orio Nacional}
\centerline{Rua General Jos\'e Cristino, 77, 20921-400 Rio de Janeiro - RJ,
Brazil}

\bigskip\bigskip\bigskip
\centerline{\bf Abstract}

\medskip
\noindent
We introduce two novel evolutionary formulations of the problem of coloring the
nodes of a graph. The first formulation is based on the relationship that exists
between a graph's chromatic number and its acyclic orientations. It views such
orientations as individuals and evolves them with the aid of evolutionary
operators that are very heavily based on the structure of the graph and its
acyclic orientations. The second formulation, unlike the first one, does not
tackle one graph at a time, but rather aims at evolving a ``program'' to color
all graphs belonging to a class whose members all have the same number of nodes
and other common attributes. The heuristics that result from these formulations
have been tested on some of the Second DIMACS Implementation Challenge benchmark
graphs, and have been found to be competitive when compared to the several other
heuristics that have also been tested on those graphs.

\bigskip\bigskip
\noindent
{\bf Keywords:} Graph coloring, evolutionary algorithms, genetic algorithms,
genetic programming.

\vfill
\noindent
{\bf Address for correspondence and proofs:}

\medskip
Valmir C. Barbosa, {\tt valmir@cos.ufrj.br}

Programa de Sistemas, COPPE/UFRJ

Caixa Postal 68511

21945-970 Rio de Janeiro - RJ

Brazil

\eject
\bigbeginsection 1. Introduction

We consider a graph $G=(N,E)$, where $N$ is the graph's set of nodes and $E$ its
set of edges, each edge being an unordered pair of nodes, called
{\it neighbors\/} of each other. Graph $G$ is said to be an {\it undirected\/}
graph, since the unordered nature of its edges assigns no ``directionality'' to
them. We let $n=\vert N\vert$ and $m=\vert E\vert$.

A {\it coloring\/} of $G$ is an assignment of {\it colors\/} (positive integers)
to the nodes in such a way that every node is assigned exactly one color and,
furthermore, no two neighbors are assigned the same color. The {\it graph
coloring problem\/} is the problem of providing $G$ with a coloring that employs
the least possible number of colors. This number is called the {\it chromatic
number\/} of $G$, denoted by $\chi(G)$.

The reasons why the graph coloring problem is important are twofold. First,
there are several areas of practical interest in which the ability to color
an undirected graph with as small a number of colors as possible has direct
influence on how efficiently a certain target problem can be solved. Such areas
include time tabling and  scheduling [12, 29], frequency assignment for use of
the electromagnetic spectrum [19], register allocation in compilers [9], printed
circuit board testing [21], and the solution of sparse linear systems
originating from finite-element meshes [35].

The other reason is that the graph coloring problem has been shown to be
computationally hard at a variety of levels: not only is its decision-problem
variant NP-complete [20, 27], but also it is NP-hard to solve it approximately
within $n^{1/7-\epsilon}$ of the optimum for any $\epsilon>0$ [1, 2, 6]. What
this means is that, given the current expectation of how likely it is for an
algorithm to be found to solve an NP-hard problem efficiently, coloring the
nodes of $G$ with at most $n^{1/7-\epsilon}\chi(G)$ colors is an intractable
problem for any $\epsilon>0$. There is a sense, then, in which graph coloring
stands among the hardest of the hard combinatorial optimization problems.

These two reasons are important enough to justify the quest for heuristics to
solve the graph coloring problem, and to place it together with the favorite
problems on which new meta-heuristics are tested. Many such approaches have
been proposed (e.g., [25] and the references therein), but we single one out
which, although it misses the graph's chromatic number by rather a wide margin
in several cases, has an important inspirational role in part of this paper.
This heuristic is based on the nodes' {\it degrees\/} (numbers of neighbors) and
on the known property that $\chi(G)\le\Delta+1$, where $\Delta$ is the maximum
degree over all nodes [7]. The heuristic visits nodes in nonincreasing order of
the so-called {\it saturation degree\/} (number of colors already used to color
neighbors) and assigns the current node the smallest color that is not already
taken by one of its neighbors [8]. Symmetry is broken by using a nonincreasing
order of degrees to which only neighbors that have not yet been colored
contribute, and after that randomly. This heuristic is known as the
{\it DSatur\/} heuristic.

Despite the problem's importance and the many approaches that have been tried
to tackle it, it was not until relatively recently that a concentrated effort
was undertaken to test several heuristics on a common set of graphs of various
sizes and characteristics. Such an effort was part of the Second DIMACS
Implementation Challenge [26], which remains to date the most comprehensive
project that we know of to have addressed a systematic experimental evaluation
of heuristics for the graph coloring problem.

Comprehensive though this effort was, only one of the approaches to make it to
the final meeting was of an evolutionary nature [16]. This approach involved the
combination of evolutionary and other search techniques, and was based on
previous work by the same authors [15]. In fact, to our knowledge the situation
remains that no purely evolutionary approach seems to have had success on the
graph coloring problem, although there have been evolutionary approaches for
restricted versions of the problem (e.g., [14]) and also other hybrid approaches
with evolutionary ingredients (e.g., [18] and the references therein). In our
view, the central difficulty is the difficulty of formulating the problem
adequately so that individuals and the appropriate evolutionary operators can be
properly identified. Suppose, for example, that we choose to approach the
formulation in the obvious, straightforward manner, and say that an individual
is simply a particular coloring. That is sound enough, but how does one go
about, say, generating a new individual from two parent colorings? Coloring some
nodes according to one parent and the others according to the other parent might
work, but not without some additional rule to handle the cases in which the same
color ends up assigned to neighbors, or else allowing such infeasible color
assignments while somehow penalizing them through appropriately low fitness
values.

As we perceive it, the problem with this na\"\i ve first approach and others
closely related to it is the same that prevents massive parallelism from working
satisfactorily on the graph coloring problem under certain other
meta-heuristics [5]: the problem speaks of a global property of graphs (how many
colors overall?), and it is a nontrivial matter to express this global property
in terms of more localized indicators, e.g., the chromatic number of
{\it subgraphs\/} of $G$ (a subgraph of $G$ is a graph whose node set is a
subset of $N$, with the corresponding subset of $E$ for edge set). In
particular, if $G_1$ and $G_2$ are distinct subgraphs of $G$, then it does not
follow that $\chi(G)=\chi(G_1)+\chi(G_2)$.

Notwithstanding such difficulties, graph coloring is a research area in which
much of the underlying structure has already been uncovered and considerable
heuristic experience has been accumulated. In this paper, we take advantage of
some key aspects of this structure and knowledge to introduce two new
evolutionary formulations of the graph coloring problem. The first formulation
belongs in the class of techniques we normally think of as genetic algorithms
[24, 33], while the second one can be thought of as an instance of genetic
programming [3, 28].

Our first approach, described in Section 2, is based on a view of the
colorings of $G$ that relates them to the {\it acyclic orientations\/}
of $G$. An acyclic orientation of $G$ is any of the possible ways of
assigning directions to the (undirected) edges of $G$ in such a way that no
{\it directed cycle\/} is formed (that is, in such a way that, by traversing
edges only according to the assigned directions it is impossible to reach the
same node more than once). In general, an orientation of $G$ can be regarded
as a function $\omega:E\to N$ such that, for $e=(i,j)\in E$, $\omega(e)=j$ if
and only if edge $e$ is directed from node $i$ to node $j$.

Let a {\it directed path\/} in $G$ according to some acyclic orientation be a
sequence of nodes, each of which is a neighbor of its predecessor, if it has
one, such that the last node can be reached from the first by traversing the
edges that connect consecutive nodes to each other along the directions assigned
to them by the orientation. The essence of our first approach is that, to any
coloring that uses $c$ colors overall there corresponds an acyclic orientation
whose longest directed path has no more than $c$ nodes; and that, conversely, to
any acyclic orientation whose longest directed path has $l$ nodes there
corresponds a coloring by no more than $l$ colors. If we let $\Omega(G)$ denote
the set of all acyclic orientations of $G$, then this gives rise to the
following relation between $\chi(G)$ and $\Omega(G)$. For $\omega\in\Omega(G)$,
we let $P(\omega)$ denote the set of all directed paths in $G$ according to
$\omega$. For $p\in P(\omega)$, let $l(p)$ be the number of nodes in $p$. Then,
from [13], we have
$$
\chi(G)=\min_{\omega\in\Omega(G)}\max_{p\in P(\omega)}l(p).
\eqno(1)
$$
Our first approach regards $\Omega(G)$ as the search space out of which
populations are formed, that is, each acyclic orientation of $G$ is an
individual. The fittest possible individual is the one whose longest directed
path is shortest among all possible individuals.

This first approach works on fixed $G$, that is, every new graph that comes
along to be colored triggers the evolution of acyclic orientations in order to
find one of high fitness. The second approach, which we describe in Section 3, 
is by contrast targeted at an entire class ${\cal G}$ of undirected graphs
sharing certain characteristics, as for example the value of $n$. It is inspired
by the DSatur heuristic discussed above, in the sense that it approaches the
coloring of $G$'s nodes as the search for a degree-based ordering of those nodes
that will yield a good coloring if nodes are colored as dictated by that
ordering.

More specifically, suppose for the moment that the number of nodes is the sole
characteristic shared by the members of ${\cal G}$, and let $\Kappa(n)$ stand
for the set of all $n!$ permutations of the sequence
$\langle 1,\ldots,n\rangle$. A member $\kappa=\langle k_1,\ldots,k_n\rangle$ of
$\Kappa(n)$ can be interpreted as the following sequence of instructions to
assign colors to the nodes of any graph $G\in{\cal G}$: first assign color
$1$ to the node having the $k_1$th largest degree, then assign to the node
having the $k_2$th largest degree the smallest color that does not conflict with
the colors possibly already assigned to its neighbors, and so on. If we for now
disregard the details that come from the existence of nodes having the same
degree, then $\Kappa(n)$ is the search space that contains the populations of
our second evolutionary approach. Fitness is in this case an average taken over
a pre-selected subclass of ${\cal G}$. An individual fitter than another will
require a smaller average number of colors over all members of that subclass
than the other.

An illustration of the essentials of the two approaches is given in Figure 1,
where $G$ is shown with $n=6$ and directions assigned to the undirected edges
in such a way that the graph's orientation is acyclic. The acyclic orientation
is meaningful in the context of our first approach only, and implies that $G$
can be colored by no more than four colors, since this is the number of nodes in
any longest directed path (e.g., the one from $a$ to $d$ through $b$ and $c$).
In fact, assigning color $1$ to node $d$, color $2$ to nodes $c$ and $e$, color
$3$ to nodes $b$ and $f$, and color $4$ to node $a$ is a legitimate coloring of
the nodes of $G$, although it is possible to do better easily by employing two
colors only. In the latter case, what follows from (1) is that another acyclic
orientation of $G$ exists for which the longest directed path has two nodes.
Readily, by reversing the directions of all edges directed away from nodes $b$
and $f$ in the figure, one gets such an orientation.

Figure 1 can also be used to illustrate the second approach, but disregarding
the directions assigned to edges altogether. Suppose that we have
$\kappa=\langle 3, 4, 5, 6, 1, 2\rangle$; that is, $\kappa$ instructs us to
assign the smallest color that does not cause conflicts to nodes $a$, $b$, $d$,
and $e$ first (this is a possible arrangement of the nodes having the third
through sixth largest degrees), for example, then to nodes $c$ and $f$ (a
possible arrangement of the nodes having the first and second largest degrees),
again for example. Clearly, four colors would be needed overall. Once again, it
is simple to see that an alternative member of $\Kappa(6)$, for example
$\kappa'=\langle 1, 2, 3, 4, 5, 6\rangle$, would lead to a coloring of the nodes
of $G$ by a total of two colors regardless of how the nodes having the same
degree were arranged with respect to one another.

\topinsert
$$
\beginpicture
\setcoordinatesystem units <1.00cm,1.00cm>
\setplotarea x from -0.50 to 4.50, y from -0.50 to 2.50

   \put {$\bullet$} at 0.00 1.00
   \put {$\bullet$} at 1.00 2.00
   \put {$\bullet$} at 1.00 0.00
   \put {$\bullet$} at 3.00 2.00
   \put {$\bullet$} at 3.00 0.00
   \put {$\bullet$} at 4.00 1.00
   \put {$a$} at -0.50  1.00
   \put {$b$} at  1.00 -0.50
   \put {$c$} at  3.00 -0.50
   \put {$d$} at  4.50  1.00
   \put {$e$} at  3.00  2.50
   \put {$f$} at  1.00  2.50

   \arrow <0.3cm> [0.1,0.4] from 0.00 1.00 to 1.00 2.00
   \arrow <0.3cm> [0.1,0.4] from 0.00 1.00 to 1.00 0.00
   \arrow <0.3cm> [0.1,0.4] from 1.00 2.00 to 3.00 2.00
   \arrow <0.3cm> [0.1,0.4] from 1.00 2.00 to 3.00 0.00
   \arrow <0.3cm> [0.1,0.4] from 1.00 0.00 to 3.00 0.00
   \arrow <0.3cm> [0.1,0.4] from 3.00 2.00 to 4.00 1.00
   \arrow <0.3cm> [0.1,0.4] from 3.00 0.00 to 4.00 1.00

\endpicture

$$
\bigskip
\centerline{{\bf Figure 1.} $G$ oriented acyclically}
\endinsert

In addition to Sections 2 and 3, the remainder of the paper comprises three
additional sections. In Section 4, a brief discussion on how the two strategies'
evolutionary operators relate to their state spaces is presented. Section 5
contains a description of the graphs that we use in our experiments, which are
drawn from the set of test graphs employed in the DIMACS implementation
challenge. Experimental results are given in Section 6. They indicate that our
two new evolutionary formulations yield algorithms that are competitive when
compared to several others, including those developed in response to the DIMACS
challenge. Section 7 contains concluding remarks.

\bigbeginsection 2. The first formulation

Our first formulation is based on the interplay between acyclic orientations
of $G$ and colorings of its nodes. If a coloring exists that uses $c$ colors
overall, then consider the orientation obtained by directing every edge from
the node with the highest color to the one with the lowest. The resulting
orientation is clearly acyclic, and furthermore no directed path exists
having more than $c$ nodes. In the same vein, consider an acyclic orientation
whose longest directed path has $l$ nodes, and consider the nodes that are
{\it sinks\/} according to this orientation (a sink is a node whose incident
edges are all directed inward by the orientation). Now assign colors to nodes as
follows. First assign color $1$ to all sinks. Then remove all sinks from $G$
(together with all edges incident to them) and assign color $2$ to all the new
sinks that are thus formed. This removal of sinks can be repeated exactly $l$
times, and each time it is repeated the resulting sinks are assigned their
color, which conceivably can be some of the colors used previously. What results
is then a coloring that employs no more than $l$ colors.

An important consequence of this interplay is (1), which relates the chromatic
number of $G$ to its acyclic orientations. Our first formulation regards acyclic
orientations as individuals and aims at evolving an acyclic orientation whose
longest directed path is as short as possible. Once this individual is
identified, the corresponding coloring of the nodes of $G$ can be found as
indicated previously.

Before we give the details of this formulation, we pause momentarily to comment
on the cardinality of $\Omega(G)$, the set of all acyclic orientations of $G$.
While for some graphs such cardinality is known as a function of $n$ or $m$ that
can be computed easily ($2^{n-1}$ for trees, $n!$ for complete graphs, $2^n-2$
for rings), the general case requires the evaluation of the so-called
{\it chromatic polynomial\/} of $G$. For $c>0$, this polynomial, denoted by
$C_G(c)$, yields the number of distinct ways in which the nodes of $G$ can be
colored by a total of at most $c$ colors [7]. Clearly, the smallest such $c$
for which $C_G(c)>0$ is equal to $\chi(G)$.

Once the chromatic polynomial of $G$ has been introduced, the cardinality of
$\Omega(G)$ can be proven to be given by
$$
\bigl\vert\Omega(G)\bigr\vert=(-1)^nC_G(-1),
$$
as demonstrated in [37]. We think this is a remarkable property, and felt it
should be presented even though it bears no further relationship to the topic of
the paper than clarifying how to assess $\bigl\vert\Omega(G)\bigr\vert$, at
least in principle, in the general case, and perhaps providing additional
evidence that colorings of $G$'s nodes and acyclic orientations of $G$ are
indeed deeply related to one another.

The remainder of this section contains a detailed description of the key
elements of our first approach. These are the assessment of an acyclic
orientation's fitness, and the functioning of the two evolutionary operators
we employ, namely crossover and mutation. We use the following additional
notation. For $\omega\in\Omega(G)$, we let $S^+(\omega)$ be the set of sinks in
$G$ according to $\omega$. Likewise, $S^-(\omega)$ denotes the set of
{\it sources\/} in $G$ according to $\omega$ (these are nodes whose incident
edges are all directed outward by $\omega$). For $i\in N$, $N_i^-(\omega)$ is
the set of neighbors $j$ of $i$ such that edge $(i,j)$ is directed outward from
$i$ by $\omega$.

\bigskip
\noindent
{\bf Fitness evaluation.} The fitness of an individual $\omega\in\Omega(G)$ is
the nonnegative integer $f_1(\omega)$, given by
$$
f_1(\omega)=n-\max_{p\in P(\omega)}l(p),
\eqno(2)
$$
where we recall that $P(\omega)$ is the set of all directed paths in $G$
according to $\omega$ and $l(p)$ is the number of nodes in directed path $p$.
It follows from (2) that an individual has higher fitness than another if its
lengthiest directed path is shorter than the other's. By our preceding
discussion, an individual $\omega$ yields a coloring of $G$'s nodes by no more
than $n-f_1(\omega)$ colors.

Clearly, an equivalent reformulation of $f_1(\omega)$ is
$$
f_1(\omega)=n-\max_{i\in S^-(\omega)}l_i,
\eqno(3)
$$
where $l_i$ is the number of nodes on the longest directed path in $G$
according to $\omega$ that starts at node $i$. The expression for $f_1(\omega)$
in (3) is more useful because the value of $l_i$ for all $i\in N$ can be
computed efficiently by straightforward depth-first search [10], as in procedure
${\it dfs}(i,\omega)$, given next.

For all $i\in N$, let ${\it reached}_i$ be a Boolean variable used to indicate
whether node $i$ has been reached by the depth-first search (it is set to
{\bf false} initially). We then have:

\bigskip
\item{} {\bf procedure} ${\it dfs}(i,\omega)$
\itemitem{} {\bf if} ${\it reached}_i={\bf false}$
\itemitemitem{} ${\it reached}_i:={\bf true}$
\itemitemitem{} {\bf if} $i\in S^+(\omega)$
\itemitemitemitem{} $l_i:=1$
\itemitemitem{} {\bf else}
\itemitemitemitem{} $l_i:=1+\max_{j\in N_i^-(\omega)}{\it dfs}(j,\omega)$
\itemitem{} Return $l_i$

\bigskip
This procedure allows us to rewrite $f_1(\omega)$ yet again as
$$
f_1(\omega)=n-\max_{i\in S^-(\omega)}{\it dfs}(i,\omega),
\eqno(4)
$$
which indicates directly how the fitness of an individual is to be evaluated
in practice, requiring for such $O(m)$ time.

\bigskip
\noindent
{\bf Crossover.} The crossover of two parent individuals
$\omega_1,\omega_2\in\Omega(G)$ to yield the two offspring
$\omega'_1,\omega'_2\in\Omega(G)$ is better understood if we adopt a linear
representation for the acyclic orientations of $G$. For $\omega\in\Omega(G)$,
say that a node $j$ is {\it reachable\/} from node $i$ according to $\omega$
if a directed path exists from $i$ to $j$. Now let the sequence of nodes
$$
L(\omega)=\langle i_1,\ldots,i_n\rangle
$$
be such that, for $1\le x,y\le n$, $x<y$ if $i_y$ is reachable from $i_x$.

If we regard $\omega$ as a partial order of the nodes of $G$, then clearly
$L(\omega)$ is a linear extension of that partial order and has, as such, the
following characteristic. For $i_x\in N$, every other node $i_y$ such that $x<y$
(that is, to the right of $i_x$ in $L(\omega)$) is either reachable from $i_x$
or neither node is reachable from the other. Note that the placement of $i_x$
and $i_y$ in $L(\omega)$ is somewhat arbitrary when no directed path exists
between the two nodes, so some symmetry-breaking rule is necessary to construct
the linear representation. When this happens, our choice is to let $x<y$ if and
only if $i_x<i_y$, where, as in sections to come, we assume for simplicity that
the set of nodes $N$ can be regarded as the set of natural numbers
$\{1,\ldots,n\}$, so that the comparison ``$i_x<i_y$'' is meaningful. In Section
6, we comment briefly on the possible effects of using randomness to break
symmetry.

We describe the crossover of $\omega_1$ and $\omega_2$ in terms of their
linear representations, namely
$$
L(\omega_1)=\langle i_1,\ldots,i_n\rangle
$$
and
$$
L(\omega_2)=\langle j_1,\ldots,j_n\rangle.
$$
If $z$ such that $1\le z<n$ is the crossover point and we let
$$
L(\omega'_1)=\langle i'_1,\ldots,i'_n\rangle
$$
and
$$
L(\omega'_2)=\langle j'_1,\ldots,j'_n\rangle,
$$
then $\langle i'_1,\ldots,i'_z\rangle=\langle i_1,\ldots,i_z\rangle$ and
$\langle i'_{z+1},\ldots,i'_n\rangle$ is the subsequence of
$\langle j_1,\ldots,j_n\rangle$ comprising all the nodes that are not already
in $\langle i_1,\ldots,i_z\rangle$. The construction of $L(\omega'_2)$ is
entirely analogous:
$\langle j'_1,\ldots,j'_z\rangle=\langle j_1,\ldots,j_z\rangle$ and
$\langle j'_{z+1},\ldots,j'_n\rangle$ is the subsequence of
$\langle i_1,\ldots,i_n\rangle$ comprising all the nodes that are not already
in $\langle j_1,\ldots,j_z\rangle$.

In order to give an interpretation of this crossover operation that is
meaningful in the context of the acyclic orientations of $G$, we must first
recognize that the linear representation we adopted is unambiguous. In fact,
any sequence $L$ comprising all nodes from $N$ can be seen to give rise to a
unique acyclic orientation, as follows. For $x<y$, if $i_x$ and $i_y$ are
neighbors in $G$, then direct the edge between them from $i_x$ to $i_y$. Any
directed cycle in the resulting orientation would imply the reflexivity of $<$,
which does not hold. Consequently, we are justified in having written
$L(\omega'_1)$ and $L(\omega'_2)$, because the unique orientations that result
from those sequences are indeed acyclic.

Not only this, but we can now see clearly what is inherited by $\omega'_1$ and
$\omega'_2$ from $\omega_1$ and $\omega_2$. Orientation $\omega'_1$ inherits
from $\omega_1$ the directions assigned to all edges joining nodes in the set
$\{i'_1,\ldots,i'_z\}$ to any other nodes, and from $\omega_2$ those assigned to
the edges whose end nodes all lie in $\{i'_{z+1},\ldots,i'_n\}$. As for
$\omega'_2$, it inherits from $\omega_2$ the directions assigned to all edges
joining nodes in the set $\{j'_1,\ldots,j'_z\}$ to any other nodes, and from
$\omega_1$ those assigned to the edges whose end nodes are all members of
$\{j'_{z+1},\ldots,j'_n\}$.

\bigskip
\noindent
{\bf Mutation.} There are several possibilities for mutation on an acyclic
orientation of $G$, but we have decided to opt for what seems to be the most
elementary one while guaranteeing the generation of another acyclic
orientation. Given an acyclic orientation $\omega$ and the mutation point
$i\in N$, the operation consists of turning node $i$ into a source to yield a
new orientation $\omega'$. That $\omega'$ is also acyclic has been argued
elsewhere (e.g., [4]), and the argument goes as follows. If $\omega'$ is not
acyclic, then the directed cycle that exists in it must involve node $i$,
which is where the only change was effected on $\omega$ to yield $\omega'$.
But this is clearly impossible, since according to $\omega'$ node $i$ is a
source.

\bigbeginsection 3. The second formulation

Our second evolutionary formulation of the graph coloring problem is
conceptually simpler than the first. The individuals that it evolves are
permutations of the sequence $\langle 1,\ldots,n\rangle$, in which each number
is used as an index into a certain reference sequence of the nodes of $G$. We
let $D=\langle i_1,\ldots,i_n\rangle$ be such a sequence, which is arranged in
nonincreasing order of node degrees and, like in Section 2, regards nodes as
natural numbers for the purpose of breaking symmetry when more than one node
with the same degree exist. In $D$, $i_x$ comes to the left of $i_y$ if either
the degree of $i_x$ is larger than the degree of $i_y$ or the two degrees are
the same but $i_x<i_y$.

Thus, recalling that $\Kappa(n)$ stands for the set of all permutations of
$\langle 1,\ldots,n\rangle$, a permutation
$\kappa=\langle k_1,\ldots,k_n\rangle\in\Kappa(n)$ is to be interpreted as a
sequence of $n$ steps to assign colors to the nodes of $G$. If we let
${\it node}(k)$ denote the $k$th node in sequence $D$ (i.e.,
${\it node}(k)=i_k$), then the $z$th step in $\kappa$ assigns to
${\it node}(k_z)$ the smallest color that has not been assigned to any of its
neighbors in $G$. Of course, the outcome of such a sequence of steps depends on
our choice on how to break symmetry; in Section 6, we comment on the possible
effects of randomness and of choosing an altogether different reference
sequence.

Each $\kappa\in\Kappa(n)$ can in principle be regarded as a {\it program\/} to
provide any graph on $n$ nodes with a coloring (hence our earlier indication
that this second formulation relates to genetic programming, even though
individuals are in our case considerably less complex in structural terms). The
goal is to evolve one such program that, on average, can use as small a number
of colors as possible. However, the number of nonisomorphic graphs on $n$ nodes
grows so fast with $n$ (over $165$ billion graphs for $n$ as small as $12$ [31])
that it is not reasonable to expect acceptable performance over such a wide
class of graphs of a program evolved within tolerable time bounds. On the other
hand, it also seems improper to proceed as with the first formulation and evolve
a program that is specific to a single graph, since now the very nature of our
individuals calls for greater generality.

Our choice has been to evolve programs that are specific to graphs belonging
to a certain class of graphs, all having in common not only the number of nodes
but also the {\it density}, which is defined to be the ratio of the graph's
number of edges to the maximum possible number of edges when there can only
be at most one edge between any two nodes. We let $p$ denote such density,
which is then such that
$$
p={{2m}\over{n(n-1)}}.
$$
The class of graphs having $n$ nodes and density $p$ such that $p^-\le p\le p^+$
for $0\le p^-\le p^+\le 1$ will be denoted by ${\cal G}(n,p^-,p^+)$.

We now describe how to assess the fitness of an individual, and how to
perform crossover and do mutation and inversion.

\bigskip
\noindent
{\bf Fitness evaluation.} Let ${\cal T}$, henceforth referred to as the
{\it training set}, be a randomly selected subset of ${\cal G}(n,p^-,p^+)$. The
fitness of an individual $\kappa\in\Kappa(n)$ is based on the average number of
colors that individual requires over all members of ${\cal T}$. If for program
$\kappa$ we let ${\it colors}(\kappa,G)$ denote the number of colors required
for $\kappa$ to assign colors to the nodes of graph $G$, then the fitness of
$\kappa$ is the nonnegative rational $f_2(\kappa)$, given by
$$
f_2(\kappa)=n-{{1}\over{\vert{\cal T}\vert}}
\sum_{G\in{\cal T}}{\it colors}(\kappa,G),
\eqno(5)
$$
which can be computed in $O \bigl(\sum_{G\in{\cal T}}m_G\bigr)$ time, where
$m_G$ denotes the number of edges in $G\in{\cal T}$. Program $\kappa$ is capable
of coloring the nodes of all graphs in ${\cal T}$ with, on average,
$n-f_2(\kappa)$ colors.

\bigskip
\noindent
{\bf Crossover.} The crossover of two parent programs
$\kappa_1,\kappa_2\in\Kappa(n)$ to yield the two offspring
$\kappa'_1,\kappa'_2\in\Kappa(n)$, given the crossover point $z$ such that
$1\le z<n$, works as follows. Let
$$
\kappa_1=\langle k_1,\ldots,k_n\rangle,
$$
$$
\kappa_2=\langle \ell_1,\ldots,\ell_n\rangle,
$$
$$
\kappa'_1=\langle k'_1,\ldots,k'_n\rangle,
$$
and
$$
\kappa'_2=\langle \ell'_1,\ldots,\ell'_n\rangle.
$$
Then $\langle k'_1,\ldots,k'_z\rangle=\langle k_1,\ldots,k_z\rangle$ and
$\langle k'_{z+1},\ldots,k'_n\rangle$ is the subsequence of
$\langle \ell_1,\ldots,\ell_n\rangle$ comprising all the indices that are not
already in $\langle k_1,\ldots,k_z\rangle$. As for $\kappa'_2$,
$\langle \ell'_1,\ldots,\ell'_z\rangle=\langle \ell_1,\ldots,\ell_z\rangle$ and
$\langle \ell'_{z+1},\ldots,\ell'_n\rangle$ is the subsequence of
$\langle k_1,\ldots,k_n\rangle$ comprising all the indices that are not already
in $\langle \ell_1,\ldots,\ell_z\rangle$.

Inheritance in this case is easier to identify than in the formulation of
Section 2. Program $\kappa'_1$ visits ${\it node}(k'_1),\ldots,{\it node}(k'_z)$
for coloring in the same relative order as program $\kappa_1$ and
${\it node}(k'_{z+1}),\ldots,{\it node}(k'_n)$ in the same relative order as
$\kappa_2$. Likewise, program $\kappa'_2$ visits
${\it node}(\ell'_1),\ldots,{\it node}(\ell'_z)$ for coloring in the same
relative order as program $\kappa_2$ and
${\it node}(\ell'_{z+1}),\ldots,{\it node}(\ell'_n)$ in the same relative order
as $\kappa_1$.

\bigskip
\noindent
{\bf Mutation.} Single-locus mutations are meaningless on a program
$\kappa\in\Kappa(n)$, so what we call a mutation in this second formulation is
really a swap of two of the indices in $\kappa$, which then acquire each other's
positions in the sequence. For mutation points $z$ and $z'$, the effect of this
operation on $\kappa=\langle k_1,\ldots,k_n\rangle$ is to yield the program
$\kappa'=\langle k'_1,\ldots,k'_n\rangle$ with $k'_z=k_{z'}$ and $k'_{z'}=k_z$.

\bigskip
\noindent
{\bf Inversion.} For program $\kappa=\langle k_1,\ldots,k_n\rangle$ in
$\Kappa(n)$, the effect of inversion on $\kappa$ to yield another program
$\kappa'=\langle k'_1,\ldots,k'_n\rangle$ is to set
$k'_z=k_{n-z+1}$ for $z=1,\ldots,n$.

\bigbeginsection 4. On the evolutionary operators

The evolutionary operators described in Sections 2 and 3 for use respectively
in our first and second formulations of the graph coloring problem were selected
with a variety of goals in mind, including meaningfulness in the context of the
formulation at hand, parsimony in the final number of operators, and expected
effectiveness (as far as this can be inferred from a few initial experiments on
reasonably-sized instances of the problem). While it is possible to see
relatively easily, as we argued in those sections, that the crossover operator
is indeed in both formulations responsible for the transmission to offspring of
structurally meaningful information from the parents, for the remaining
operators (mutation and, in the case of the second formulation, inversion as
well) it may seem that they only fulfill the elusive role of occasionally
helping the search escape local optima [17, 32]. In this section, we argue that
mutation, in both formulations, can be ascribed the more precise role of
allowing the search to lead from any one individual to any other with nonzero
probability (possibly with the help of inversion, in the case of the second
formulation). Understanding this property, which holds trivially in the case of
bit-string representations under single-locus mutations, for example, requires a
little elaboration in the case of our two formulations, especially so for the
first one.

We first discuss mutation in the context of Section 2. In this case, mutation
on an acyclic orientation is the turning of an arbitrary node (the mutation
point) into a source, which yields another acyclic orientation. For arbitrary
$\omega,\omega'\in\Omega(G)$, the question that we answer affirmatively in this
section is whether there exists a finite sequence of mutations that turns
$\omega$ into $\omega'$. The argument requires a little additional notation, as
follows. Let $M\subseteq E$ be the set of edges that $\omega$ and $\omega'$
direct differently, and let $U\subseteq N$ be the set of end nodes of the edges
in $M$. Likewise, let $M^+\subseteq E$ be the set of all edges lying on directed
paths according to $\omega$ that lead to nodes in $U$, and let $U^+\subseteq N$
be the set of end nodes of the edges in $M^+$. Clearly, $U\subseteq U^+$ and
$M\subseteq M^+$. The desired transformation of $\omega$ into $\omega'$ must
reverse the directions assigned by $\omega$ to all edges in $M$ while keeping
all the remaining edges' directions untouched.

In order to describe a finite sequence of mutations that achieves this we need
still more notation. For $k\in N$, let $t_k$ be a nonnegative integer indicating
the number of times node $k$ must be turned into a source in such a sequence.
We show that there exist an assignment of values to such integers and a
corresponding sequence of mutations that always guarantee the transformation of
$\omega$ into $\omega'$. In this assignment, we first let $t_k=0$ for every $k$
such that $k\in N\setminus U^+$ (here, and henceforth, $\setminus$ is used to
denote set difference). As for $k\in U^+$, we let values be assigned in such a
way that
$$
\eqalign{
t_i=t_j,&\hbox{\quad if\ }(i,j)\notin M,\cr
t_i<t_j,&\hbox{\quad if\ }(i,j)\in M\cr
}
\eqno(6)
$$
for $(i,j)\in M^+$ directed by $\omega$ from $i$ to $j$. What (6) is doing is to
assign values that are strictly increasing as one traverses edges of $M$ along
the directions given by $\omega$, but remain constant over the traversal of
edges of $M^+\setminus M$ also along the directions given by $\omega$.

Note that such an assignment is always possible, owing to the fact that $\omega$
is acyclic, which implies that the strict inequalities appearing in (6) for
members of $M$ can always be satisfied. In particular, because values are
nondecreasing along directed paths, it is a trivial matter to choose them in
such a way that $t_k<\vert U^+\vert$ for all $k\in U^+$.

If $\{t_k; k\in U^+\}$ is a set of values that obeys (6), then the corresponding
sequence of mutations that leads from $\omega$ to $\omega'$ is obtained by
repeating the following step until $t_k=0$ for all $k\in U^+$. If $\ell\in U^+$
is not currently a source and $t_\ell$ is nonzero and greatest over all of
$U^+$, then turn $\ell$ into a source and set $t_\ell$ to $t_\ell-1$; when two
neighbors are tied, pick the one that is pointed to by the edge between them
according to the current orientation. Note that, if every $t_k$ is assigned the
smallest possible value, then this sequence of mutations comprises
$O\bigl(\vert U^+\vert^2\bigr)$ mutations overall.

Now let $e=(i,j)$ be any edge, and let the direction assigned to it by $\omega$
be from $i$ to $j$. If $e$ is not an edge of $M^+$, then clearly the process we
just described does not affect it. If $e\in M^+\setminus M$, then $t_i$ and
$t_j$ have initially the same value, and therefore $i$ and $j$ get turned into
sources the same number of times and alternately with respect to each other.
Also, because $e$ is directed toward $j$ by $\omega$, $j$ is the first of the
two nodes to be turned into a source and $i$ is the last, so at the end the
edge's direction settles as initially, that is, from $i$ to $j$. For $e\in M$,
initially it holds that $t_i<t_j$, so $j$ is turned into a source strictly more
times than $i$ is. Two scenarios can be envisaged. In the first, the inequality
$t_i<t_j$ is maintained throughout the process, so $j$ is the last of the two
nodes to be turned into a source. In the second scenario, $t_i$ and $t_j$ become
equal before the end of the process, which must happen upon one of the turnings
of $j$ into a source, therefore leaving the edge directed toward $i$. From then
on the two nodes become sources the same number of times and alternately, $i$
being the first and $j$ the last. In either scenario, edge $e$ ends up directed
from $j$ to $i$. It follows from this that $\omega'$ is the orientation of the
graph at the end of the sequence of mutations.

The case of the formulation of Section 3 is considerably simpler. For two
programs $\kappa=\langle k_1,\ldots,k_n\rangle$ and
$\kappa'=\langle k'_1,\ldots,k'_n\rangle$, the following is a sequence of
mutations that leads from $\kappa$ to $\kappa'$. For $z=1,\ldots,n$, check
whether $k'_z=k_z$; if not, then let $z'>z$ be such that $k'_z=k_{z'}$ and
swap $k_z$ with $k_{z'}$. Clearly, the resulting $\kappa$ is the same as
$\kappa'$. Also, fast though this transformation of $\kappa$ into $\kappa'$ is
(no more than $n-1$ swaps are ever needed), it is easy to conceive of cases in
which the use of inversion can make it even faster when applied as a first step.

\bigbeginsection 5. The experimental test set

This section contains a brief description of the graphs to be used in Section 6
as benchmark graphs for implementations of our first and second formulations.
They have all been extracted from the DIMACS challenge suite [38]. In what
follows, $G$ is a graph of the type being described.

\bigskip
\noindent
{\tt DSJC$n$.$q$} [25]. These are random graphs on $n$ nodes. They are
constructed by independently joining any unordered pair of nodes by an edge with
probability $q/10$. We use these graphs with $n\in\{125,250,500\}$ and $q=5$.

\bigskip
\noindent
{\tt R$n$.$q$}. These are random graphs on $n$ nodes. They are constructed
by randomly placing the nodes inside a unit square and then connecting by an
edge any two nodes that are no farther apart from each other than $q/10$. We use
these graphs with $n\in\{125,250\}$ and $q\in\{1,5\}$.

\bigskip
\noindent
{\tt R$n$.$q$c}. These graphs are the {\it complements\/} of the
{\tt R$n$.$q$} graphs, that is, each one of them has exactly the edges that are
absent from its counterpart. We use these graphs with $n\in\{125,250\}$ and
$q=1$.

\bigskip
\noindent
{\tt DSJR$n$.$q$} [25]. The same as {\tt R$n$.$q$}. We use $n=500$ and $q=1$.

\bigskip
\noindent
{\tt DSJR$n$.$q$c} [25]. The same as {\tt R$n$.$q$c}. We use $n=500$ and $q=1$.

\bigskip
\noindent
{\tt flat$n$\_$k$\_$f$} [11]. These are {\it $k$-partite\/} random graphs on $n$
nodes with density $p\approx 0.5$. Being $k$-partite means that the node set can
be partitioned into $k$ {\it independent sets}, that is, sets whose members are
not neighbors; consequently, $\chi(G)\le k$. By construction, all independent
sets have as close to the same number of nodes as possible, and furthermore
edges are distributed among pairs of independent sets as equitably as possible.
In addition, given any two of the independent sets, no node in one of them is
allowed more edges joining it to the other set than any other node in its own
set (except for a number $f$ of edges, known as a ``flatness'' parameter). We
use these graphs with $n=300$, $k\in\{20,26,28\}$, and $f=0$.

\bigskip
\noindent
{\tt le$n$\_$k$x} [29]. These are random graphs on $n$ nodes with $\chi(G)=k$.
They have numbers of edges such that $p$, the density, is such that $p\le 0.25$,
and are constructed by the creation of cliques of varying sizes in such a way
that the pre-specified value $k$ of $\chi(G)$ is not violated. We use these
graphs with $n=450$ and $k=15$; {\tt x} is one of {\tt a}, {\tt b}, {\tt c}, or
{\tt d}, and is used to differentiate among instances.

\bigskip
\noindent
{\tt mulsol.i.1} [30]. This graph results from an instance of the problem of
register allocation in compilers.

\bigskip
\noindent
{\tt school1-nsh} [30]. This graph results from the problem of constructing
class schedules in such a way as to avoid conflicts. Because it comes from an
actual class schedule, it is known that $\chi(G)\le 14$.

\bigskip
We show in Table 1 all the graphs we employ in our experiments. For each graph
$G$, the table displays the values of $n$, $m$, and $p$, as well as $\chi(G)$
(or an upper bound on it), when known from design characteristics. The table
also contains the graph class of which it is a member that is targeted in our
experiments by the implementation of our second formulation.

\topinsert
\centerline{{\bf Table 1.} Benchmark graphs}
\abovedisplayskip=0pt
$$
\vbox{
\settabs\+
&\skipbefore
&{\sc flat300\_20\_0}
&\skipbetween
&{$000$}
&\skipbetween
&{$000{,}000$}
&\skipbetween
&{$0.000$}
&\skipbetween
&{$\le 00$}
&\skipbetween
&{${\cal G}(000,0.00,0.00)$}
&\skipafter
&\cr

\+&\hrulefill
&\hrulefill
&\hrulefill
&\hrulefill
&\hrulefill
&\hrulefill
&\hrulefill
&\hrulefill
&\hrulefill
&\hrulefill
&\hrulefill
&\hrulefill
&\hrulefill
&\cr

\+&&\hfill{$G$}\hfill
&&\hfill{$n$}\hfill
&&\hfill{$m$}\hfill
&&\hfill{$p$}\hfill
&&\hfill{$\chi(G)$}\hfill
&&\hfill{Target class}\hfill
&&\cr

\vskip-\medskipamount
\+&\hrulefill
&\hrulefill
&\hrulefill
&\hrulefill
&\hrulefill
&\hrulefill
&\hrulefill
&\hrulefill
&\hrulefill
&\hrulefill
&\hrulefill
&\hrulefill
&\hrulefill
&\cr

\+&&{\tt R125.1}\hfill
&&{$125$}
&&\hfill{$209$}
&&{$0.027$}
&&\hfill{$$}
&&{${\cal G}(125,0.02,0.09)$}
&&\cr

\+&&{\tt R125.5}\hfill
&&{$125$}
&&\hfill{$3{,}838$}
&&{$0.495$}
&&\hfill{$$}
&&{${\cal G}(125,0.40,0.99)$}
&&\cr

\+&&{\tt DSJC125.5}\hfill
&&{$125$}
&&\hfill{$3{,}891$}
&&{$0.502$}
&&\hfill{$$}
&&{${\cal G}(125,0.40,0.99)$}
&&\cr

\+&&{\tt R125.1c}\hfill
&&{$125$}
&&\hfill{$7{,}501$}
&&{$0.968$}
&&\hfill{$$}
&&{${\cal G}(125,0.40,0.99)$}
&&\cr

\+&&{\tt mulsol.i.1}\hfill
&&{$197$}
&&\hfill{$3{,}925$}
&&{$0.203$}
&&\hfill{$$}
&&{${\cal G}(197,0.10,0.39)$}
&&\cr

\+&&{\tt R250.1}\hfill
&&{$250$}
&&\hfill{$867$}
&&{$0.028$}
&&\hfill{$$}
&&{${\cal G}(250,0.02,0.09)$}
&&\cr

\+&&{\tt R250.5}\hfill
&&{$250$}
&&\hfill{$14{,}849$}
&&{$0.477$}
&&\hfill{$$}
&&{${\cal G}(250,0.40,0.99)$}
&&\cr

\+&&{\tt DSJC250.5}\hfill
&&{$250$}
&&\hfill{$15{,}668$}
&&{$0.503$}
&&\hfill{$$}
&&{${\cal G}(250,0.40,0.99)$}
&&\cr

\+&&{\tt R250.1c}\hfill
&&{$250$}
&&\hfill{$30{,}227$}
&&{$0.971$}
&&\hfill{$$}
&&{${\cal G}(250,0.40,0.99)$}
&&\cr

\+&&{\tt flat300\_20\_0}\hfill
&&{$300$}
&&\hfill{$21{,}375$}
&&{$0.477$}
&&\hfill{$\le 20$}
&&{${\cal G}(300,0.40,0.99)$}
&&\cr

\+&&{\tt flat300\_26\_0}\hfill
&&{$300$}
&&\hfill{$21{,}633$}
&&{$0.482$}
&&\hfill{$\le 26$}
&&{${\cal G}(300,0.40,0.99)$}
&&\cr

\+&&{\tt flat300\_28\_0}\hfill
&&{$300$}
&&\hfill{$21{,}695$}
&&{$0.484$}
&&\hfill{$\le 28$}
&&{${\cal G}(300,0.40,0.99)$}
&&\cr

\+&&{\tt school1-nsh}\hfill
&&{$352$}
&&\hfill{$14{,}612$}
&&{$0.237$}
&&\hfill{$\le 14$}
&&{${\cal G}(352,0.10,0.39)$}
&&\cr

\+&&{\tt le450\_15a}\hfill
&&{$450$}
&&\hfill{$8{,}168$}
&&{$0.081$}
&&\hfill{$15$}
&&{${\cal G}(450,0.02,0.09)$}
&&\cr

\+&&{\tt le450\_15b}\hfill
&&{$450$}
&&\hfill{$8{,}169$}
&&{$0.081$}
&&\hfill{$15$}
&&{${\cal G}(450,0.02,0.09)$}
&&\cr

\+&&{\tt le450\_15c}\hfill
&&{$450$}
&&\hfill{$16{,}680$}
&&{$0.165$}
&&\hfill{$15$}
&&{${\cal G}(450,0.10,0.39)$}
&&\cr

\+&&{\tt le450\_15d}\hfill
&&{$450$}
&&\hfill{$16{,}750$}
&&{$0.166$}
&&\hfill{$15$}
&&{${\cal G}(450,0.10,0.39)$}
&&\cr

\+&&{\tt DSJR500.1}\hfill
&&{$500$}
&&\hfill{$3{,}555$}
&&{$0.028$}
&&\hfill{$$}
&&{${\cal G}(500,0.02,0.09)$}
&&\cr

\+&&{\tt DSJC500.5}\hfill
&&{$500$}
&&\hfill{$62{,}624$}
&&{$0.502$}
&&\hfill{$$}
&&{${\cal G}(500,0.40,0.99)$}
&&\cr

\+&&{\tt DSJR500.1c}\hfill
&&{$500$}
&&\hfill{$121{,}275$}
&&{$0.972$}
&&\hfill{$$}
&&{${\cal G}(500,0.40,0.99)$}
&&\cr

\vskip-\medskipamount
\+&\hrulefill
&\hrulefill
&\hrulefill
&\hrulefill
&\hrulefill
&\hrulefill
&\hrulefill
&\hrulefill
&\hrulefill
&\hrulefill
&\hrulefill
&\hrulefill
&\hrulefill
&\cr

}
$$
\endinsert

\bigbeginsection 6. Experimental results

Henceforth, we let {\sc Evolve\_AO} and {\sc Evolve\_P} denote the algorithms
resulting, respectively, from the evolutionary formulations of Sections 2 and 3.
{\sc Evolve\_AO} works on a fixed graph $G$ and seeks the acyclic orientation of
$G$ whose lengthiest directed path is shortest over $\Omega(G)$.
{\sc Evolve\_P}, by contrast, searches for the program (a member of $\Kappa(n)$)
that colors each of the graphs in ${\cal G}(n,p^-,p^+)$ with as few colors as
possible.

Both {\sc Evolve\_AO} and {\sc Evolve\_P} iterate for $g$ generations, each one
characterized by a population of fixed size $s$. Upon generating the last
population, each algorithm outputs the best individual found during all the
evolutionary search, denoted respectively by $\omega^*$ and $\kappa^*$. For
$k>1$, the $k$th population is generated from the $k-1$st population by first
transferring the $fs$ fittest individuals to the new population (this is an
elitist first step, with $0\le f<1$), and then repeatedly selecting individuals
for application of the evolutionary operators. The resulting individuals are
then added to the new population until it is filled. The first population is
generated randomly, which can be achieved in a similar fashion by both
algorithms by randomly creating $s$ sequences of $n$ integers. In each of these
sequences, every element of $\{1,\ldots,n\}$ must appear exactly once. For
{\sc Evolve\_AO}, one such sequence is identified with the linear representation
$L(\omega)$ of some acyclic orientation $\omega$; for {\sc Evolve\_P}, it is
identified with a program directly.

As with the choice of evolutionary operators for both algorithms, choosing an
appropriate selection method has relied on the outcome of some preliminary
experiments on reasonably-sized graphs. For {\sc Evolve\_AO}, this has resulted
in selecting individuals proportionally to its linearly normalized fitness in
the current population. In other words, if for $k\in\{1,\ldots,s\}$ a certain
individual $\omega$ has the $k$th smallest value of $f_1(\omega)$ over the
entire population for $f_1(\omega)$ as in (4), then it is selected with
probability proportional to
$$
g_1(\omega)=2k.
\eqno(7)
$$
Ties in the order of fitness within the
population are broken by the order in which individuals were added to the
population in the first place.

The case of {\sc Evolve\_P} is similar, but because fitnesses are now averages
taken over the members of a subclass of graphs ${\cal T}$ of
${\cal G}(n,p^-,p^+)$, there is a markedly increased tendency for individuals
to be clustered very nearly one another according to their fitness values.
The way we do selection in this case is then proportionally to a function that
very steeply increases with $f_2(\kappa)$ for $\kappa$ in the current
population, $f_2(\kappa)$ being given as in (5). Specifically, we let $\kappa$
be selected with probability proportional to
$$
g_2(\kappa)=e^{f_2(\kappa)}.
\eqno(8)
$$

The decision as to how to manipulate the individuals selected according to
either (7) or (8) before inclusion in the new population is also reached
probabilistically. The various probabilities involved are denoted as follows.

\medskip
\item{} $p_c$: the probability of performing the crossover of two selected
individuals;

\medskip
\item{} $p_m$: the probability of performing mutation on a selected individual;

\medskip
\item{} $p_i$: the probability of performing inversion on a selected individual.

\medskip
In this section, we summarize the results of our experiments with
{\sc Evolve\_AO} and {\sc Evolve\_P} on the benchmark graphs described in
Section 5.\footnote{$^1$}
{All experiments were performed either on a Sun UltraSparc station running
at $167$ MHz or on machines based on Intel processors running at $450$ MHz.
The time elapsed during each of the reported runs of either algorithm varies
greatly, depending on the machine used and of course on the graph, or graph
class, under consideration. Typical figures range from a few minutes to several
hours.}
The summary we present is far from exhaustive with respect to the
possible combinations of the parameters involved, but rather contains the
combinations that yielded the best results we have obtained.

We present two types of performance figures. Figures of the first type reflect
how well a graph or class of graphs is colored by a given individual. For
{\sc Evolve\_AO}, which operates on fixed $G$, this performance figure is the
number of colors needed to provide $G$ with a coloring. During the evolutionary
search, this number is given as $n-f_1(\omega^+)$, where $\omega^+$ is the best
individual of the current generation; likewise, once $\omega^*$ has been
identified as the best individual found during the whole search, then it is
given as $n-f_1(\omega^*)$. For {\sc Evolve\_P}, which aims at coloring all the
graphs belonging to the class ${\cal G}(n,p^-,p^+)$ by the program $\kappa^*$
that on average performs best on the training set
${\cal T}\subset{\cal G}(n,p^-,p^+)$, this first performance figure is presented
under two guises. During the evolutionary search, it is presented as the average
number of colors needed by the graphs in ${\cal T}$ under the best program, say
$\kappa^+$, of the current generation, that is, $n-f_2(\kappa^+)$. After
$\kappa^*$ has been identified, then the performance figure is presented as
${\it colors}(\kappa^*,G)$ for each graph $G\in{\cal G}(n,p^-,p^+)$ that is in
the benchmark set as introduced in Section 5.

Our second performance figure is related to how the quality of the best
individual output by the search is affected by the number $g$ of generations
elapsed before completion and the number $s$ of individuals of each generation.
This figure is given directly by the combined values of $g$ and $s$, and is to
be thought of as a platform-independent assessment of each algorithm's
efficiency.

This section also contains a comparative assessment of {\sc Evolve\_AO} and
{\sc Evolve\_P} with respect to seven other approaches that have also been
tested on the DIMACS benchmark graphs. These are the six approaches that
appeared in the challenge's proceedings volume [26] and the one in [18],
which appeared later. Because nearly all nine approaches are based on widely
differing strategies and have been implemented on an equally diverse set of
platforms, our comparison is restricted to performance figures that indicate
how well the benchmark graphs are colored. The following is a brief description
of those seven approaches.

\bigskip
\noindent
{\sc I\_Greedy} [11]. This is an iterated version of the greedy procedure that
assigns colors to nodes following a certain order and choosing, for each node,
the smallest color that still has not been assigned to any of its neighbors.
At each iteration, a different order is chosen according to a criterion that
involves the latest color assignment and also degree-related information, as in
DSatur, among other indicators.

\bigskip
\noindent
{\sc T\_B\&B} [23]. This is a branch-and-bound algorithm that employs elements
inspired in tabu search [22] for control. For each subproblem, it requires a
{\it clique\/} (a subgraph whose nodes are all joined to one another by edges)
of large size (number of nodes) to be found, its size being then used as a lower
bound. The algorithm is good for either exact or heuristic graph coloring,
depending on whether the large clique that is found can be guaranteed to be
maximum (have maximum size), since it can be easily shown that
$\chi(G)\ge\omega(G)$, where $\omega(G)$ is the size of the maximum clique of
$G$ [7]. Providing such a guarantee is of course as hard as coloring the graph
[20, 27], but seems to become only tolerably burdensome as the graphs become
sparser.

\bigskip
\noindent
{\sc Dist} [34]. This algorithm employs multiple sequential computers to work
concurrently on several partial colorings of the graph (that is, assignments
of colors to subsets of $N$). One of the computers maintains a pool of the best
partial colorings so far output by the others, and continually submits such
colorings to those computers to be extended by neighborhood search and possibly
improved.

\bigskip
\noindent
{\sc Par} [30]. This is a combination of {\sc Dist} with exhaustive search by
parallel branch-and-bound. The two computations are started concurrently on a
parallel machine, each one taking a certain number of processors. The processor
that in {\sc Dist} is responsible for maintaining a pool of partial colorings
is now also fed by the exhaustive-search computation, and relays the best
partial colorings found by one computation to the other.

\bigskip
\noindent
{\sc E\_B\&B} [36]. This is an exact branch-and-bound algorithm. Its key
ingredients include finding a maximum clique to work as lower bound (cf.\ our
earlier description of {\sc T\_B\&B}) and a novel branching rule that, like
DSatur, uses the nodes' saturation degrees in the decision.

\bigskip
\noindent
{\sc T\_Gen\_1} [16]. This is a hybrid approach that employs tabu search inside
a genetic algorithm. In this approach, an individual is a partition of the
graph's nodes into color classes, that is, subsets of nodes all of which are
to be assigned the same color. The generation of new individuals by crossover
does not in principle guarantee the absence of edges joining nodes in the same
color class, which is where tabu search comes in looking for the re-arrangement
into color classes that requires the least number of classes.

\bigskip
\noindent
{\sc T\_Gen\_2} [18]. This approach is entirely analogous to {\sc T\_Gen\_1},
from which it differs mainly on how crossover is performed (by ``partitioning''
color classes, as opposed to ``assigning'' colors to nodes).

\bigskip
The plots in Figure 2 show the behavior of {\sc Evolve\_AO}, each plot
corresponding to a run of the algorithm on one of the graphs in Table 1. They
have all been obtained with $p_c=0.60$, $p_m=0.02$, $f=0.70$, $g=5{,}000$, and
$s=100$. For clarity of exposition, the algorithm's behavior prior to the
hundredth generation is not shown (at the first generation, and shortly onward,
the number of colors implied by the best individual tends to be inordinately
high). Analogously, Figure 3 contains one plot for each of the eleven distinct
target classes appearing in Table 1, each plot corresponding to a run of
{\sc Evolve\_P} on randomly selected members of its target class. These plots
have been obtained with parameters valued as in Table 2.

The plots in Figures 2 and 3 indicate that both {\sc Evolve\_AO} and
{\sc Evolve\_P} exhibit the fitness-improvement behavior one expects of well
formulated evolutionary computations, often very markedly so in the earliest
generations. While for {\sc Evolve\_AO} this happens smoothly over time, for
{\sc Evolve\_P} some plots are a little jagged, perhaps reflecting the greater
difficulty of concomitantly improving the colorings of all graphs in the
training set ${\cal T}$.

\topinsert
$$
\vbox{
{
\newdimen\graphwidth\graphwidth=105pt
\newdimen\betweengraphswidth\betweengraphswidth=5pt
\newdimen\beforegraphamount\beforegraphamount=20pt
\newdimen\aftergraphamount\aftergraphamount=0pt
\def\skipgraphwidth{\hskip\graphwidth}
\def\skipbetweengraphs{\hskip\betweengraphswidth}
\def\skipbeforegraph{\vskip\beforegraphamount}
\def\skipaftergraph{\vskip\aftergraphamount}

\settabs\+
&{\skipgraphwidth}
&{\skipbetweengraphs}
&{\skipgraphwidth}
&{\skipbetweengraphs}
&{\skipgraphwidth}
&{\skipbetweengraphs}
&{\skipgraphwidth}
&\cr

\+&{\epsfxsize=\graphwidth\epsfbox{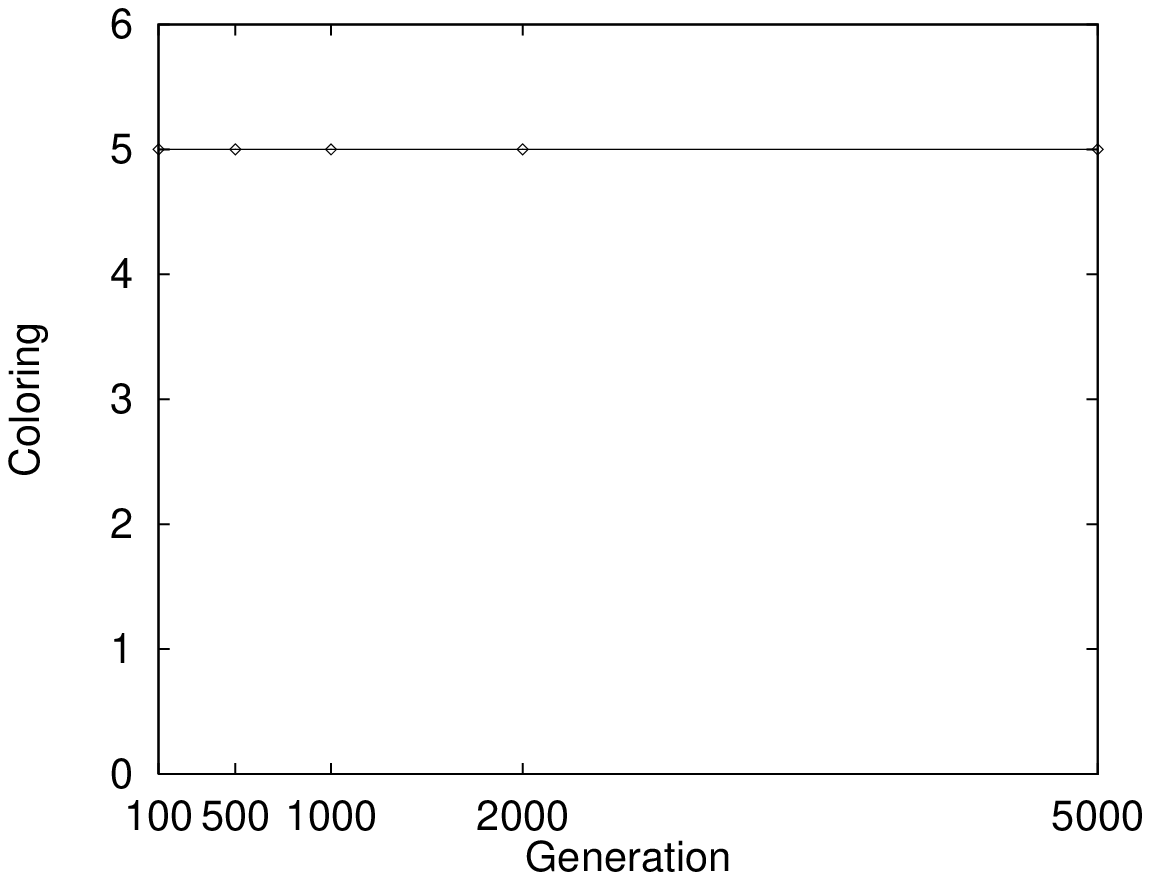}}
&&{\epsfxsize=\graphwidth\epsfbox{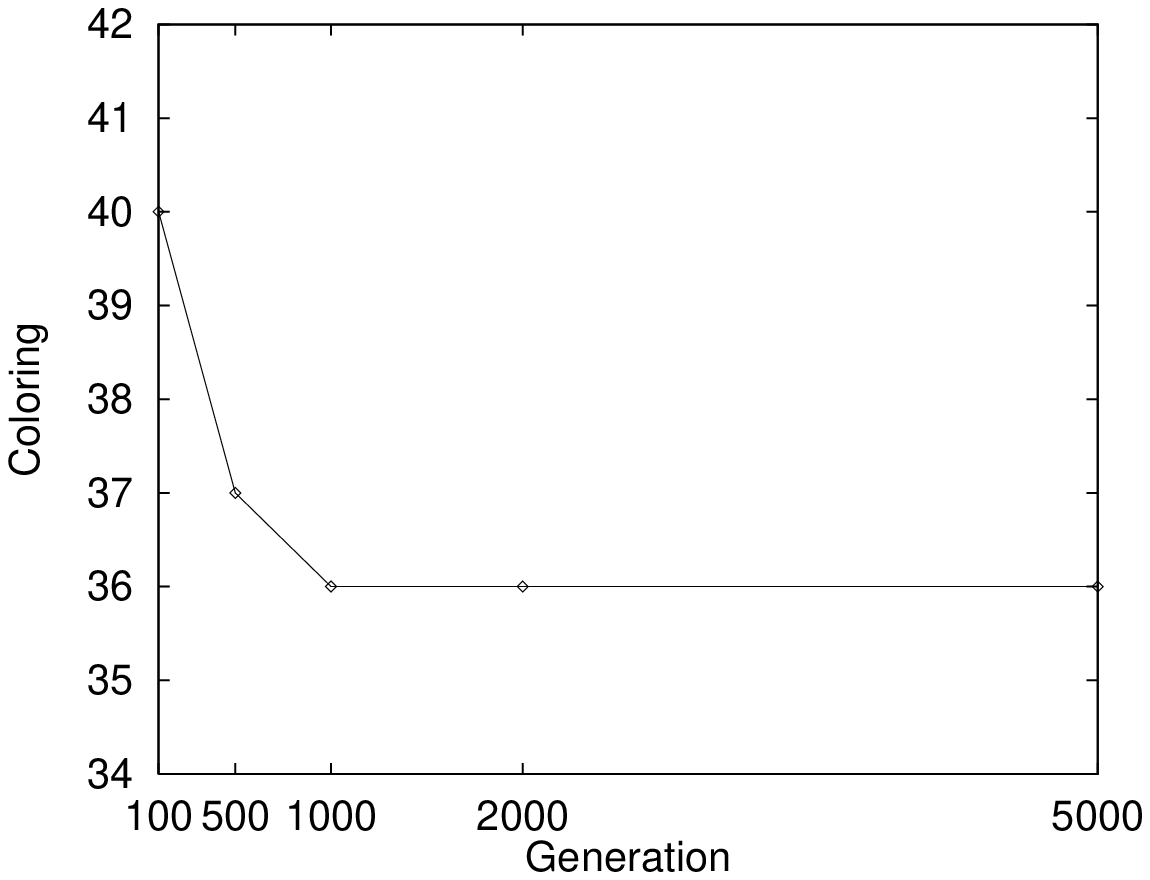}}
&&{\epsfxsize=\graphwidth\epsfbox{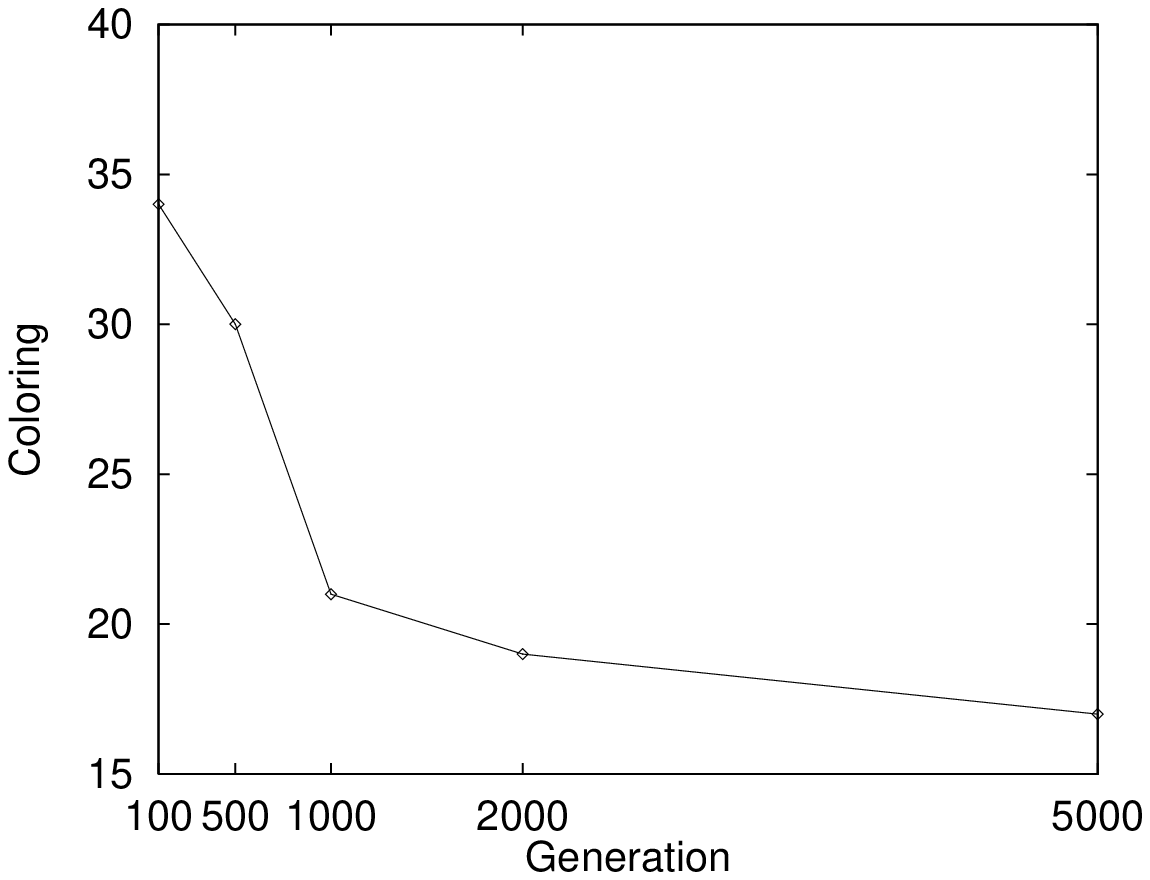}}
&&{\epsfxsize=\graphwidth\epsfbox{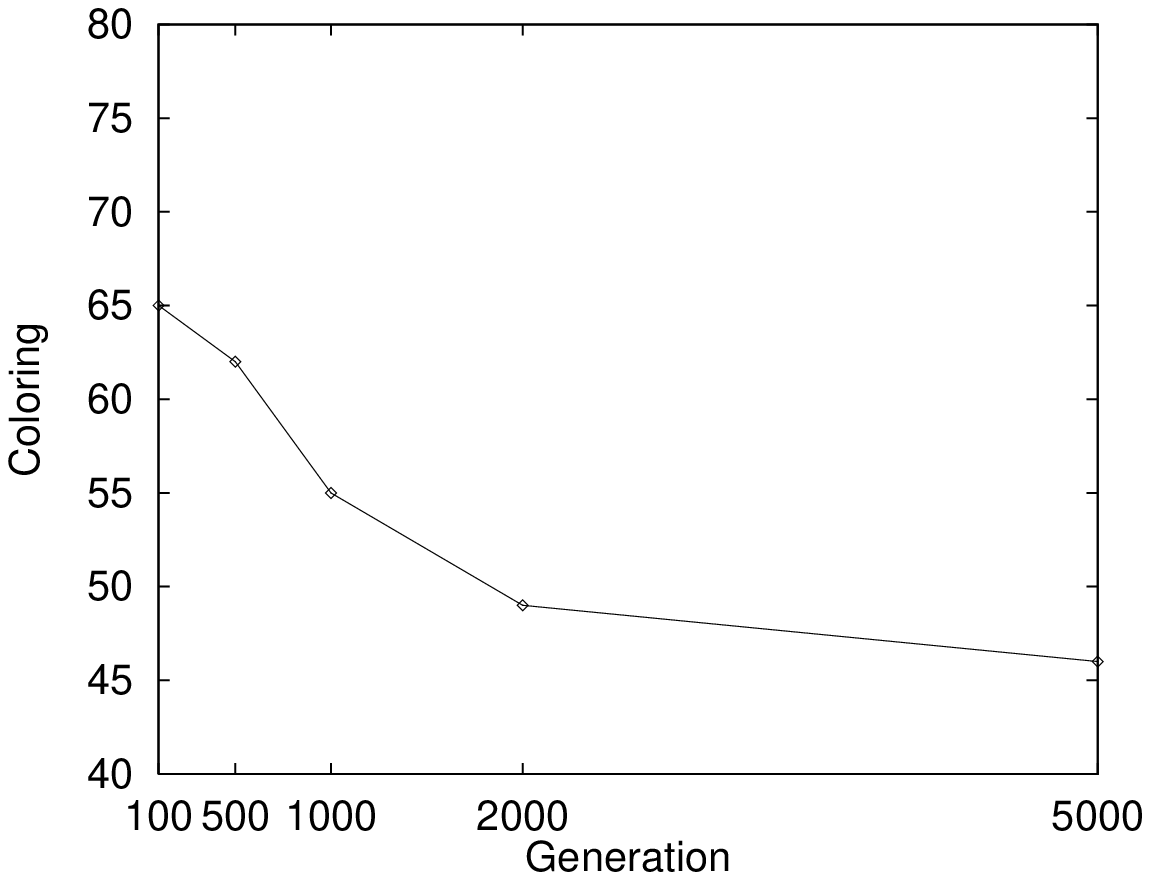}}
&\cr

\skipaftergraph
\+&\hfill{\tt R125.1}\hfill
&&\hfill{\tt R125.5}\hfill
&&\hfill{\tt DSJC125.5}\hfill
&&\hfill{\tt R125.1c}\hfill
&\cr

\skipbeforegraph
\+&{\epsfxsize=\graphwidth\epsfbox{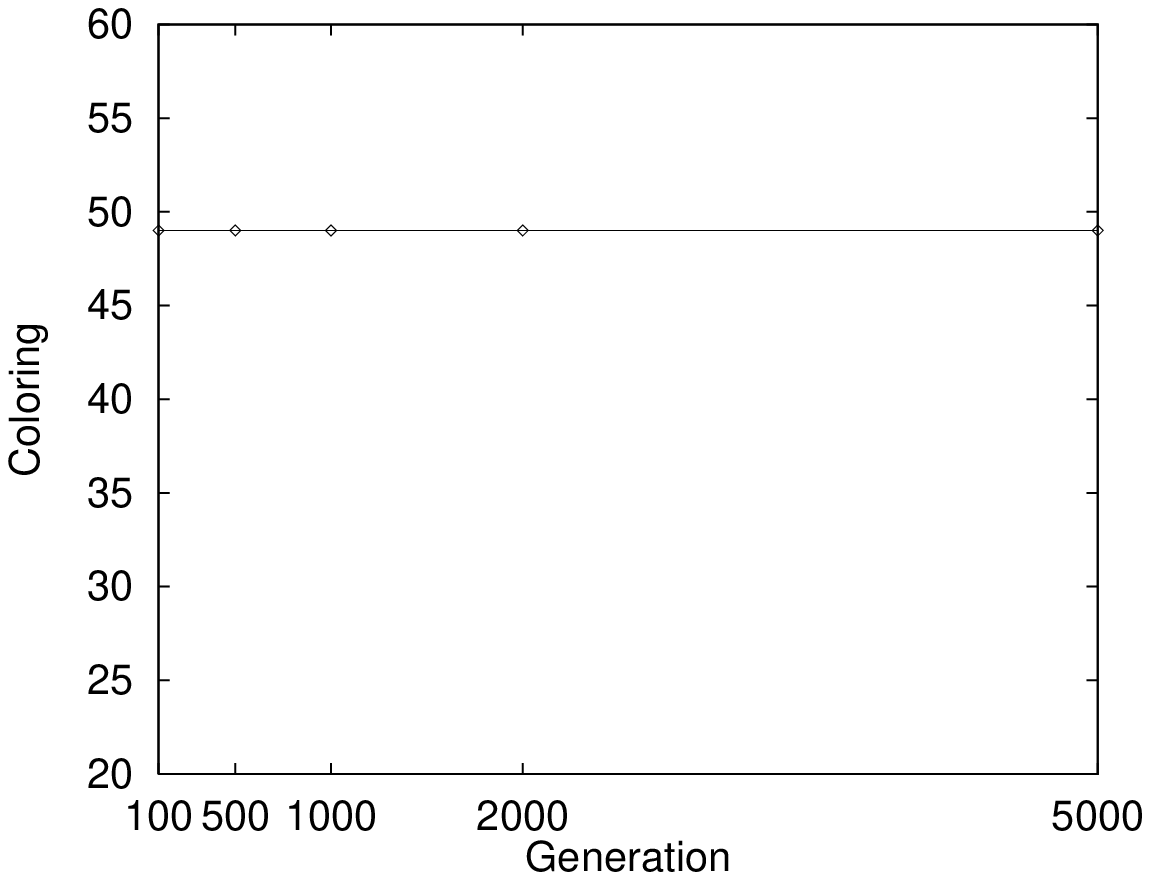}}
&&{\epsfxsize=\graphwidth\epsfbox{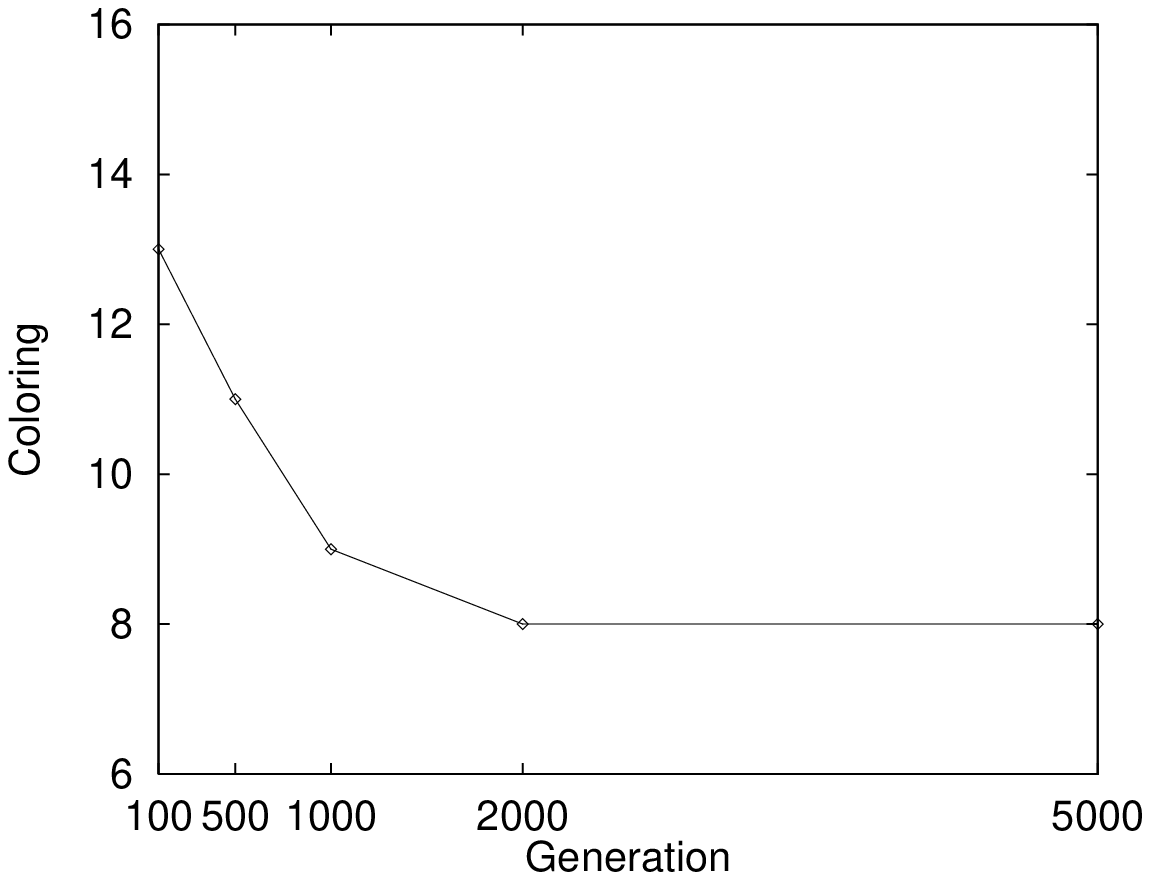}}
&&{\epsfxsize=\graphwidth\epsfbox{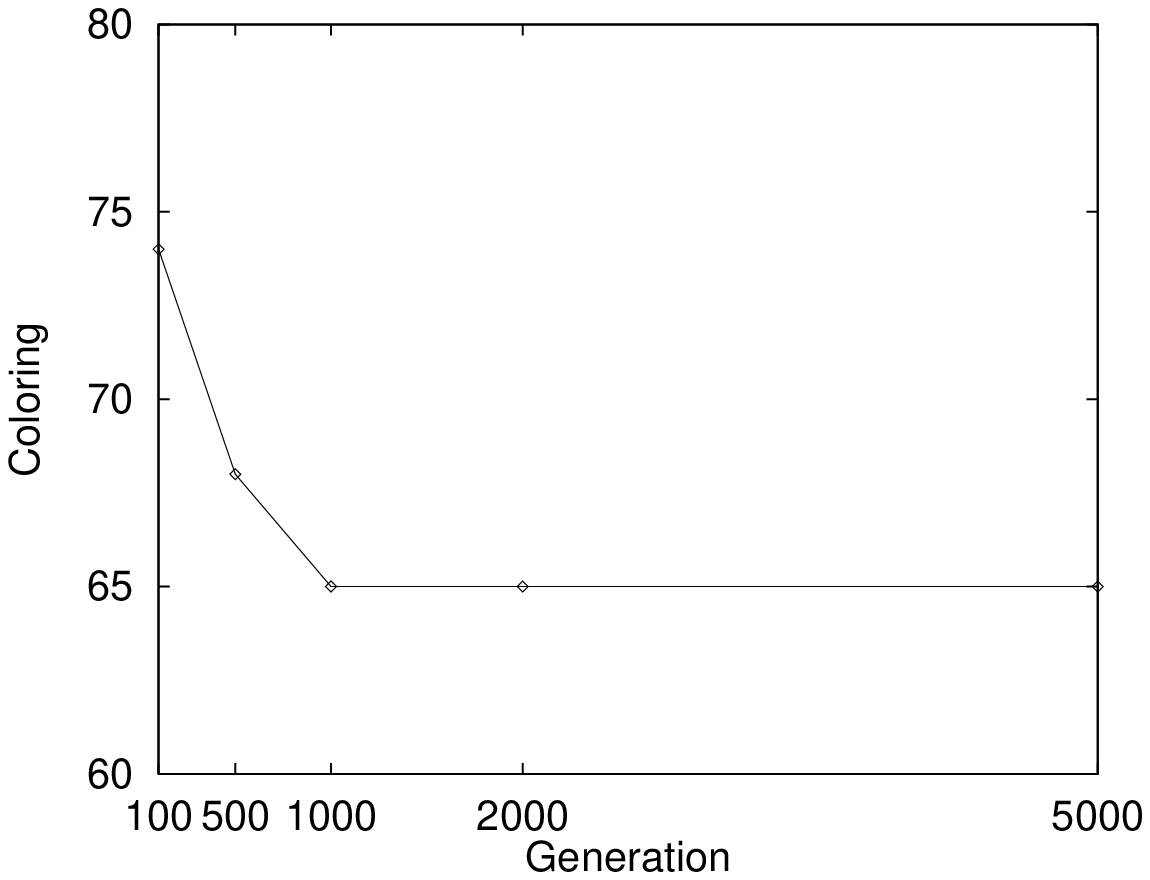}}
&&{\epsfxsize=\graphwidth\epsfbox{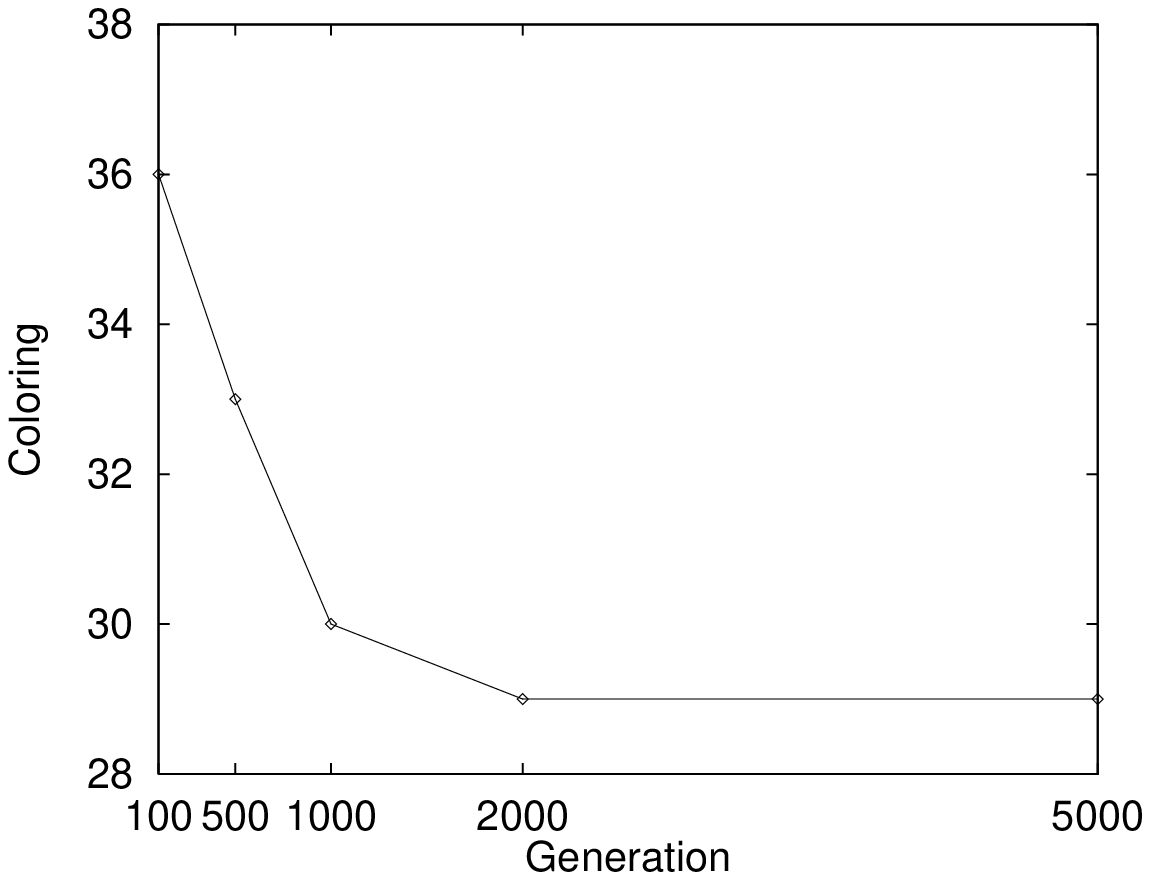}}
&\cr

\skipaftergraph
\+&\hfill{\tt mulsol.i.1}\hfill
&&\hfill{\tt R250.1}\hfill
&&\hfill{\tt R250.5}\hfill
&&\hfill{\tt DSJC250.5}\hfill
&\cr

\skipbeforegraph
\+&{\epsfxsize=\graphwidth\epsfbox{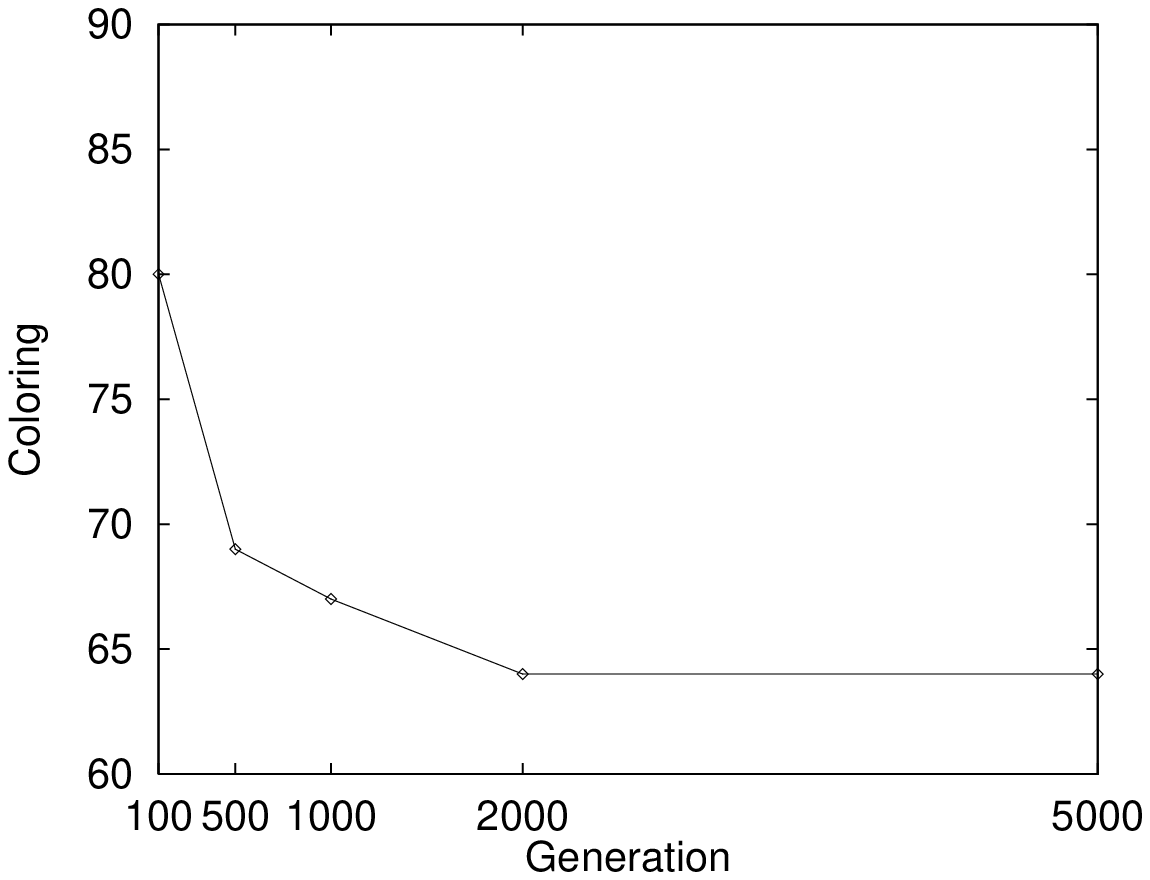}}
&&{\epsfxsize=\graphwidth\epsfbox{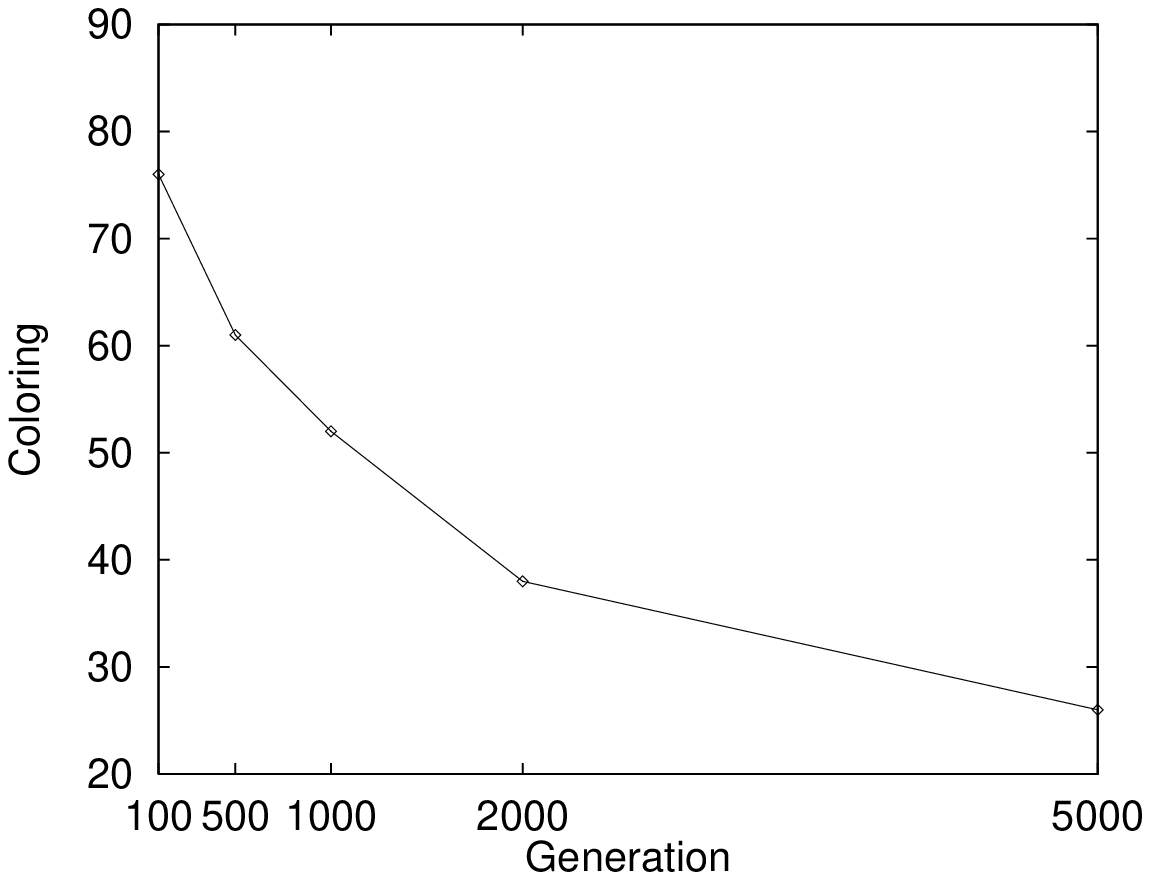}}
&&{\epsfxsize=\graphwidth\epsfbox{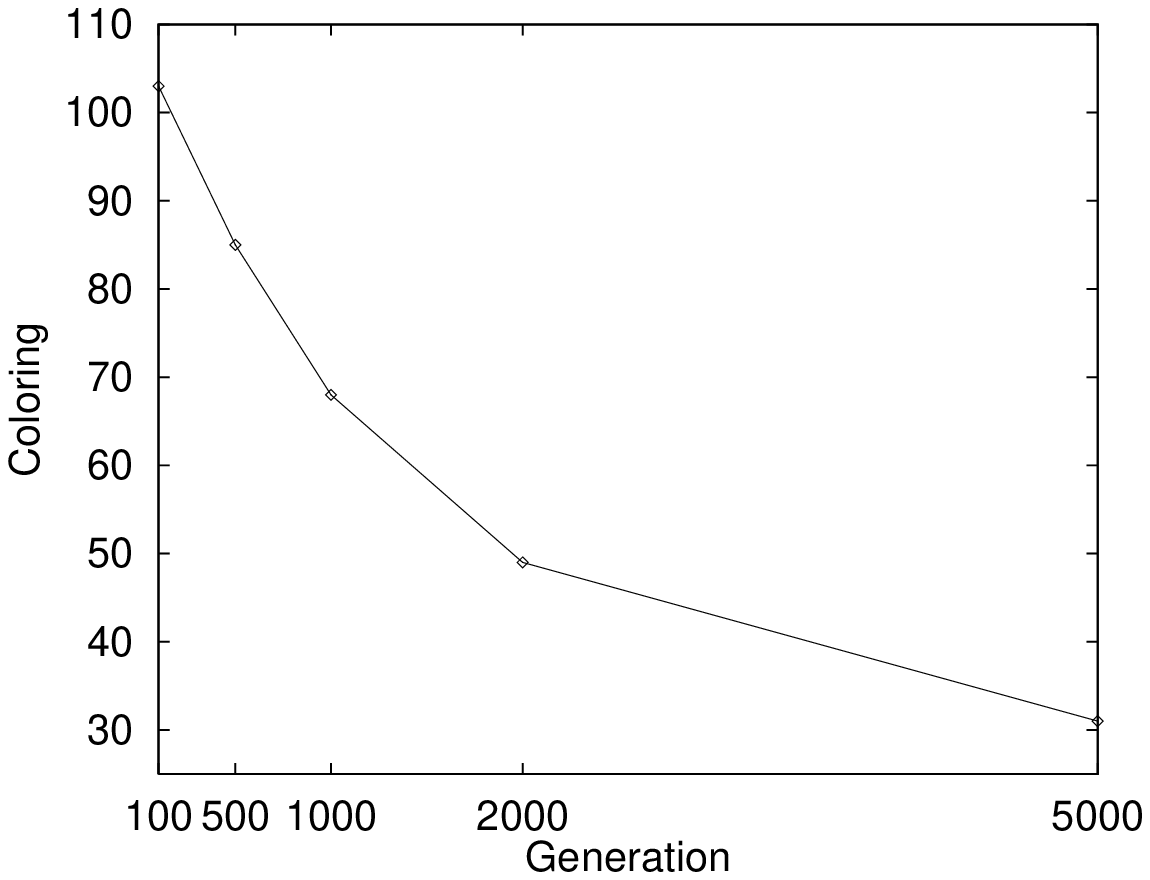}}
&&{\epsfxsize=\graphwidth\epsfbox{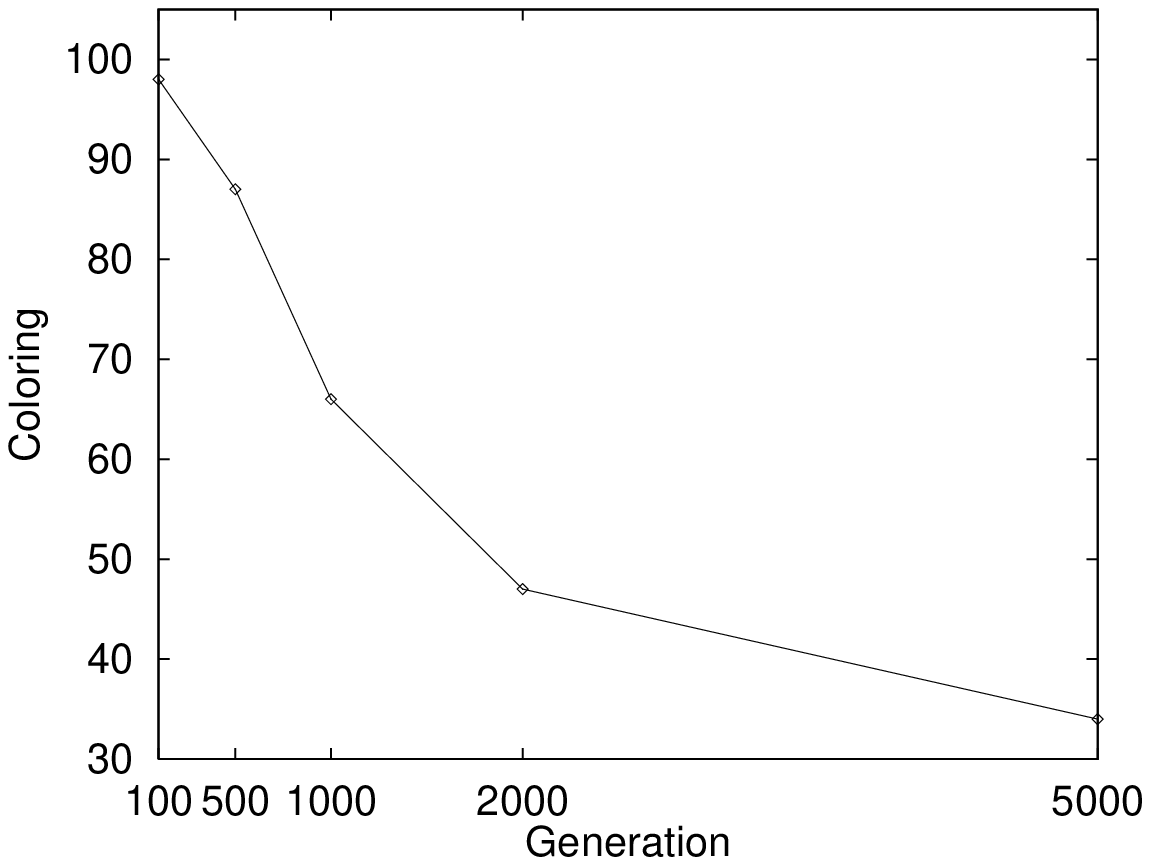}}
&\cr

\skipaftergraph
\+&\hfill{\tt R250.1c}\hfill
&&\hfill{\tt flat300\_20\_0}\hfill
&&\hfill{\tt flat300\_26\_0}\hfill
&&\hfill{\tt flat300\_28\_0}\hfill
&\cr

\skipbeforegraph
\+&{\epsfxsize=\graphwidth\epsfbox{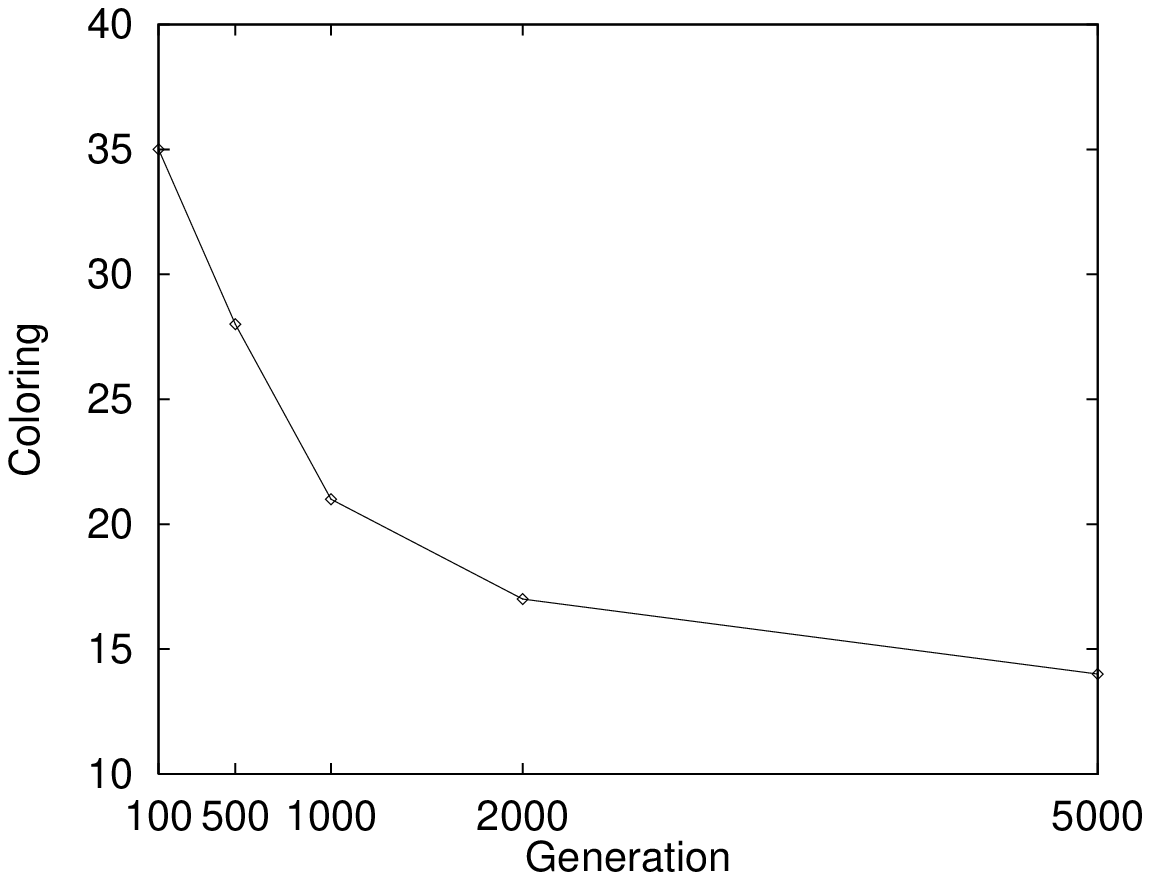}}
&&{\epsfxsize=\graphwidth\epsfbox{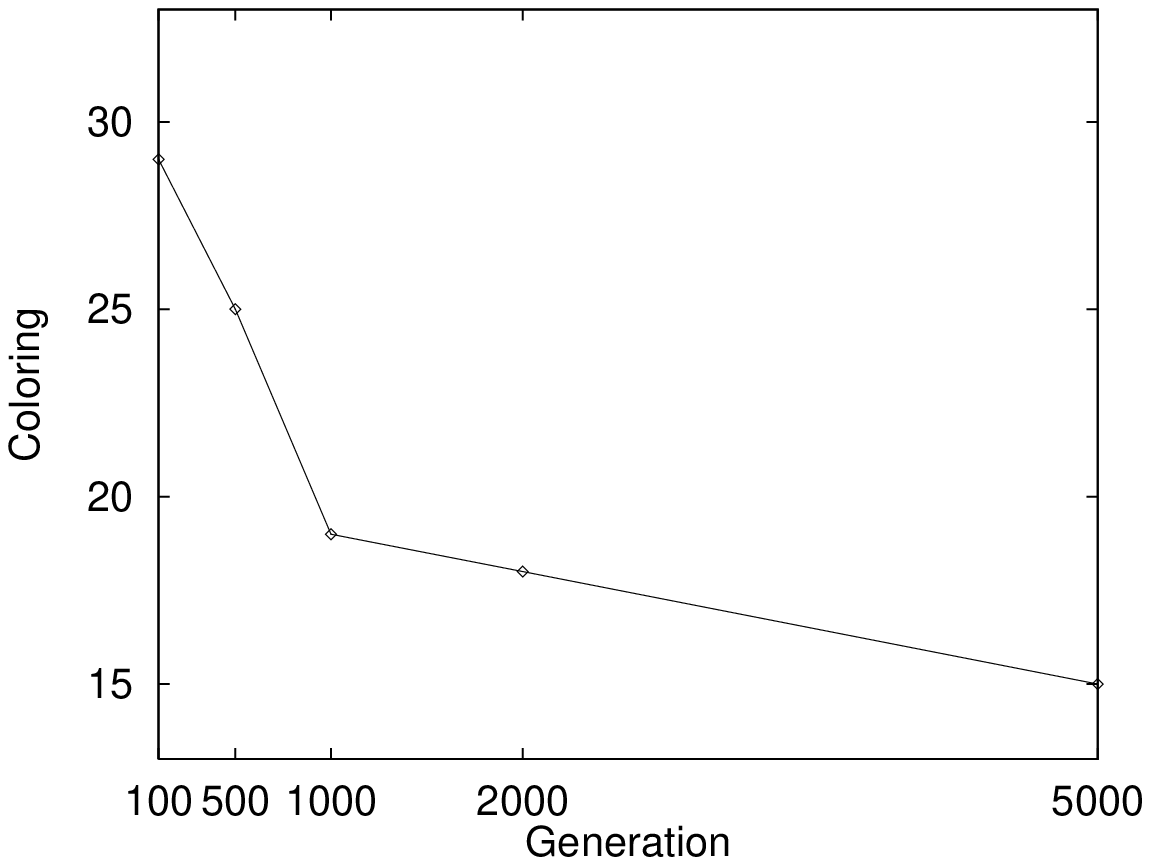}}
&&{\epsfxsize=\graphwidth\epsfbox{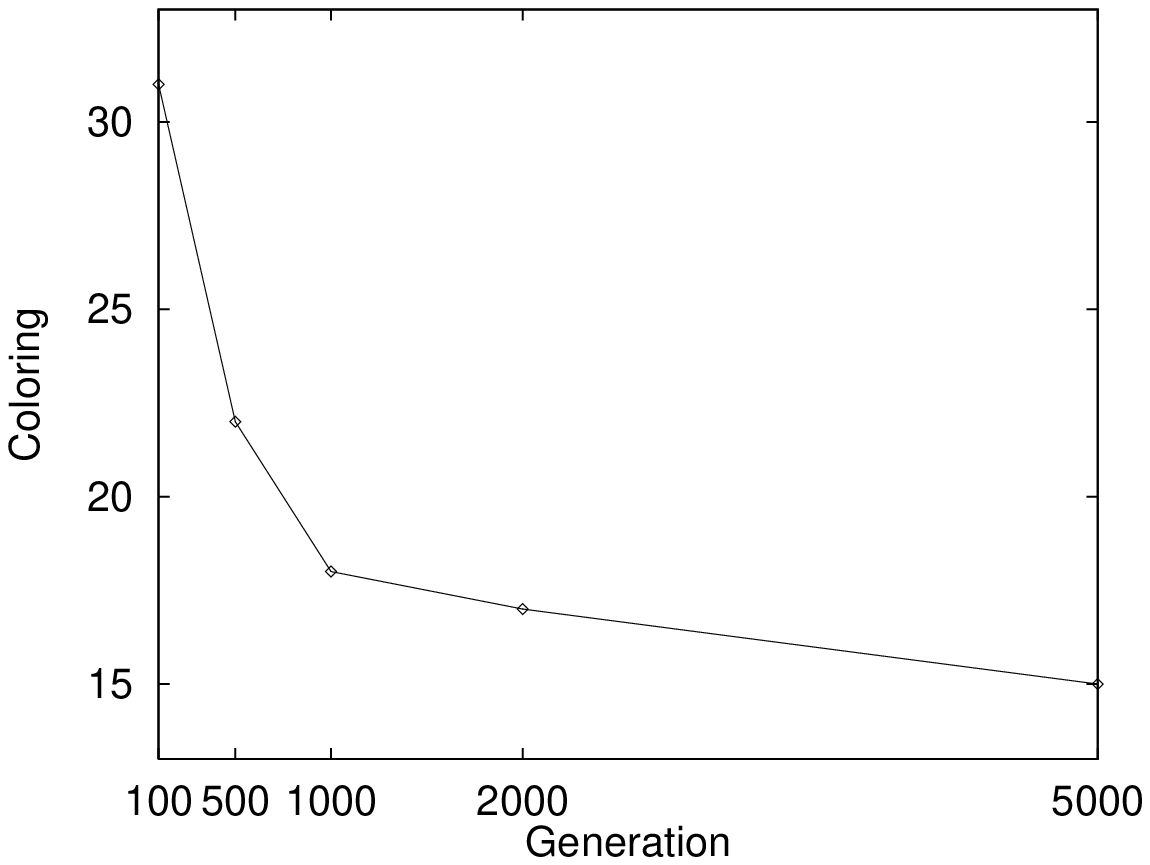}}
&&{\epsfxsize=\graphwidth\epsfbox{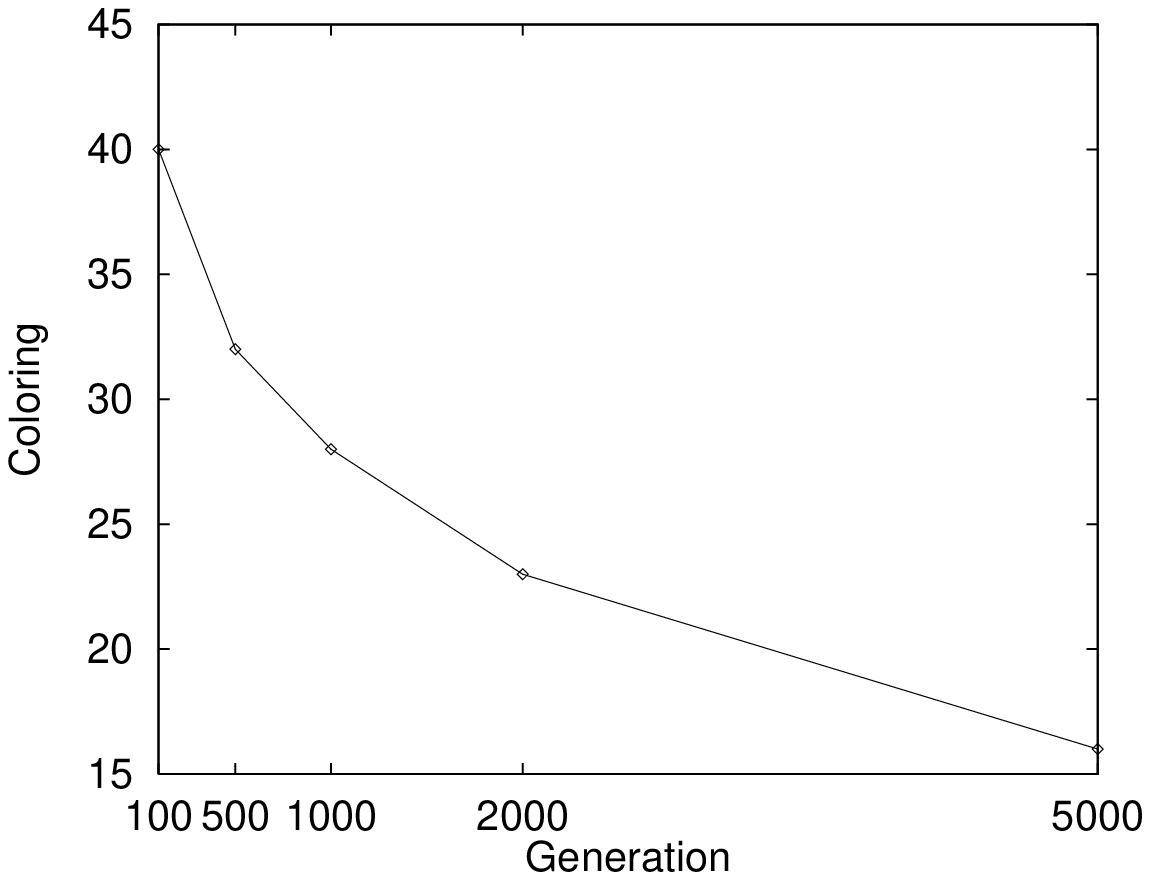}}
&\cr

\skipaftergraph
\+&\hfill{\tt school1-nsh}\hfill
&&\hfill{\tt le450\_15a}\hfill
&&\hfill{\tt le450\_15b}\hfill
&&\hfill{\tt le450\_15c}\hfill
&\cr

\skipbeforegraph
\+&{\epsfxsize=\graphwidth\epsfbox{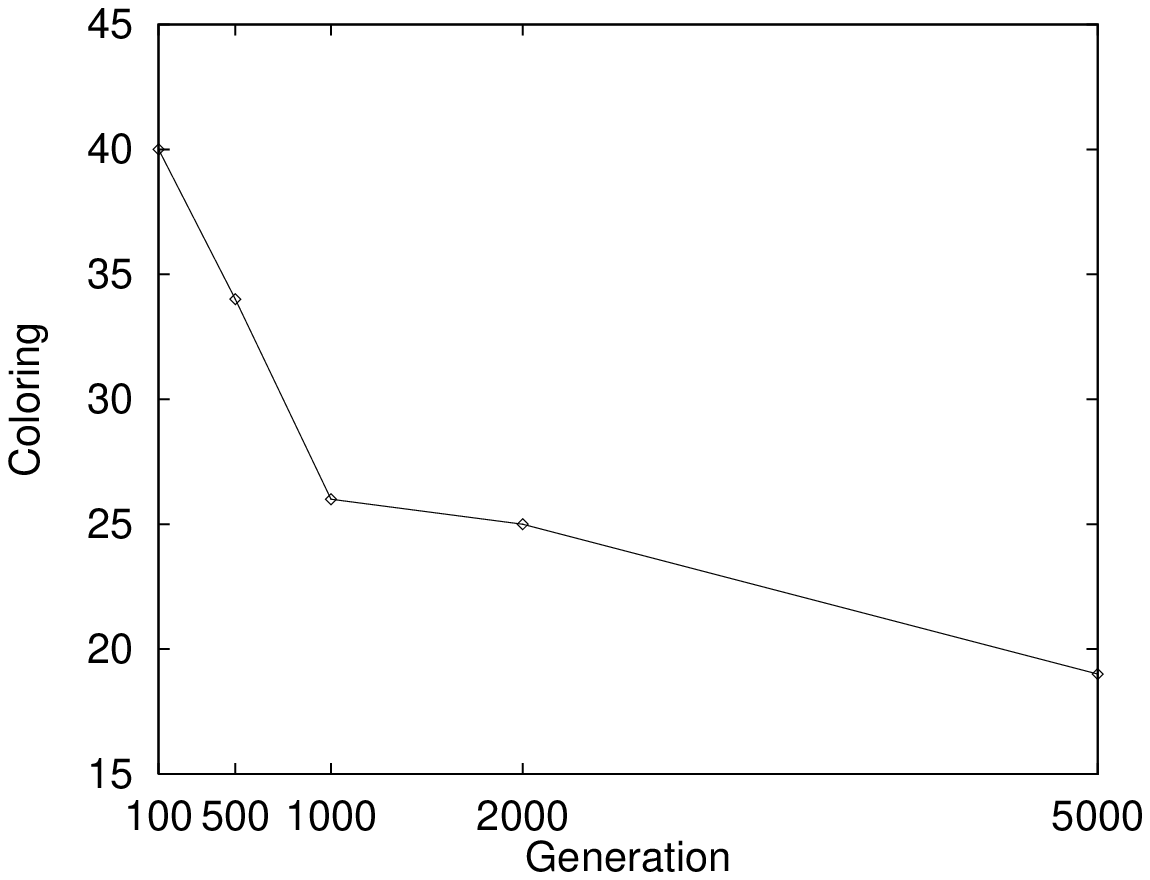}}
&&{\epsfxsize=\graphwidth\epsfbox{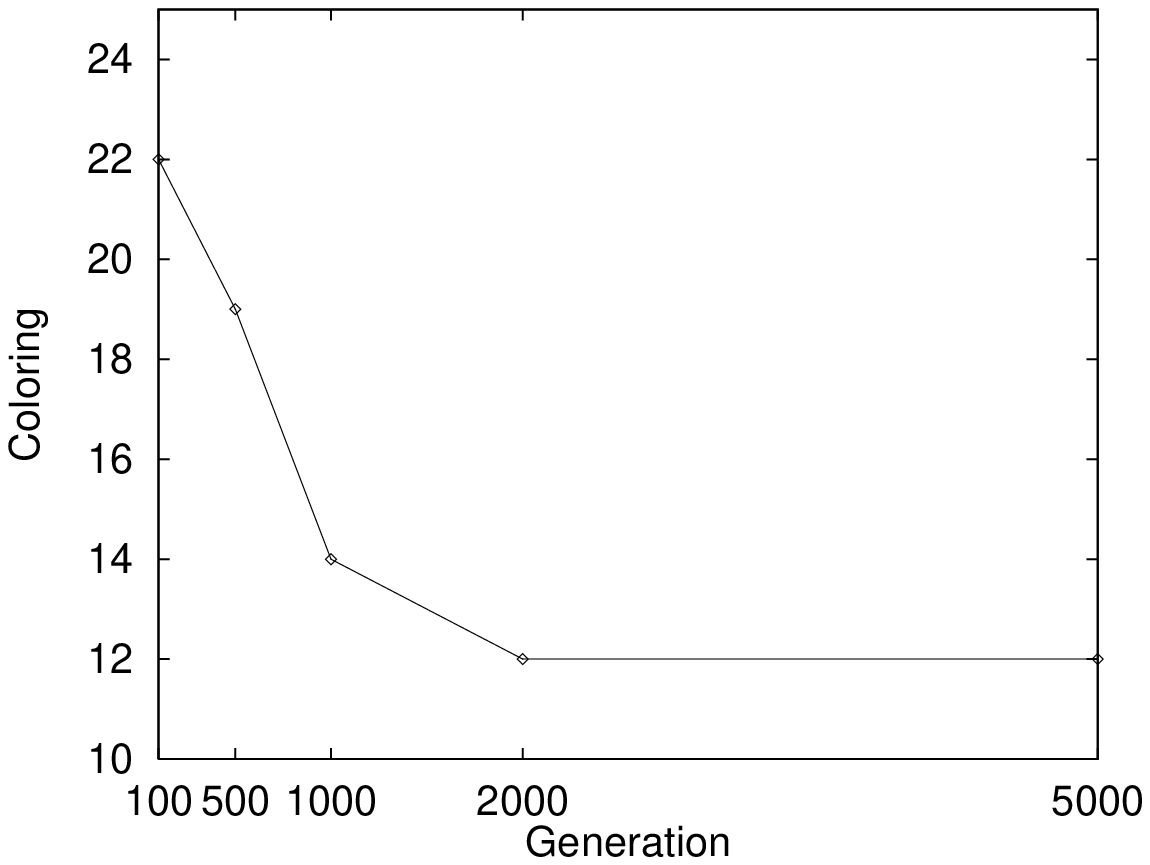}}
&&{\epsfxsize=\graphwidth\epsfbox{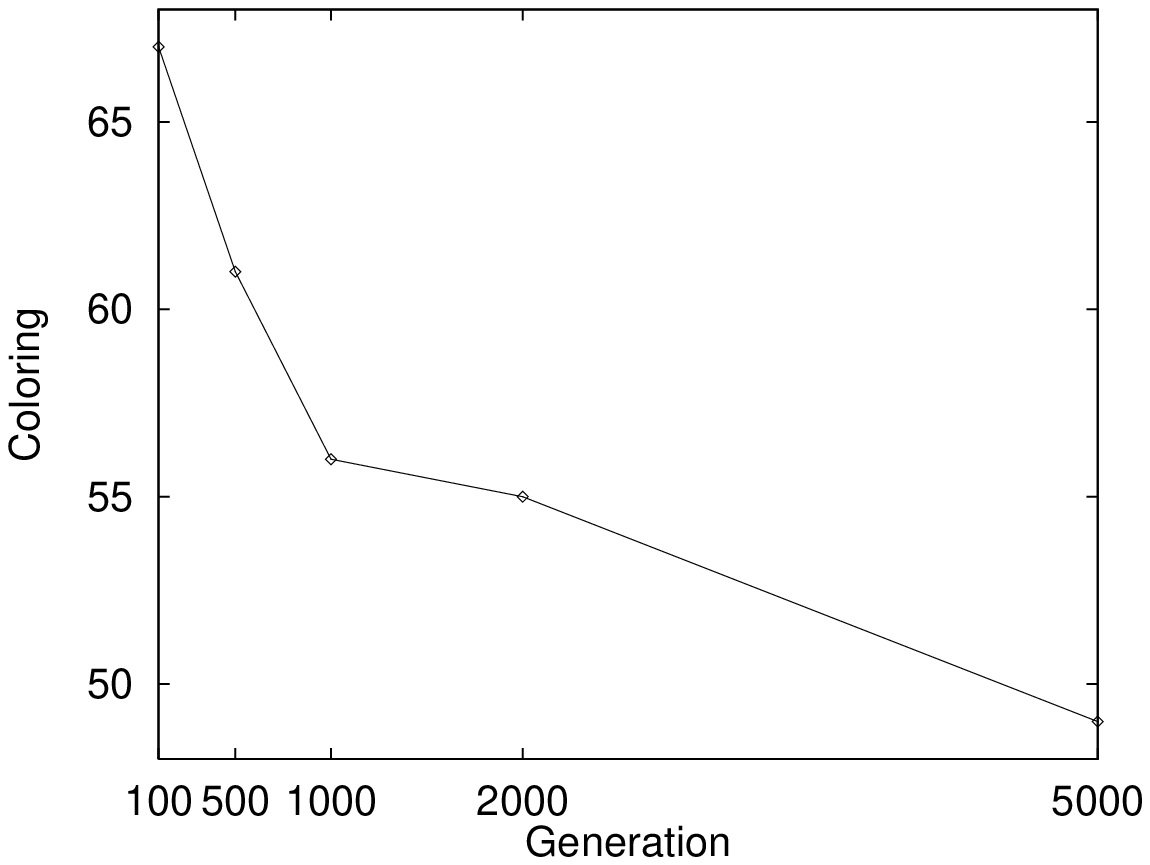}}
&&{\epsfxsize=\graphwidth\epsfbox{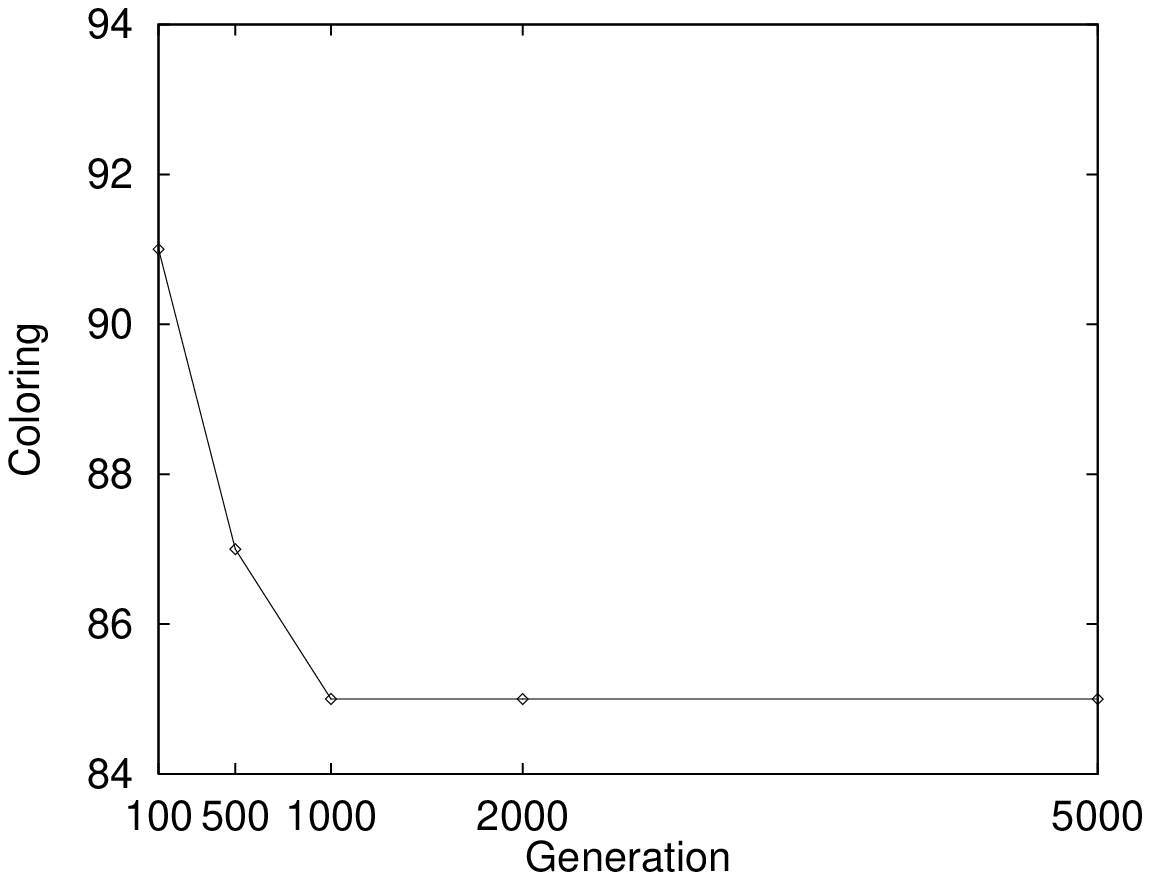}}
&\cr

\skipaftergraph
\+&\hfill{\tt le450\_15d}\hfill
&&\hfill{\tt DSJR500.1}\hfill
&&\hfill{\tt DSJC500.5}\hfill
&&\hfill{\tt DSJR500.1c}\hfill
&\cr

}

}
$$
\bigskip
\centerline{{\bf Figure 2.} Convergence of {\sc Evolve\_AO} on the graphs of
Table 1}
\endinsert

\topinsert
$$
\vbox{
{
\newdimen\graphwidth\graphwidth=135pt
\newdimen\betweengraphswidth\betweengraphswidth=10pt
\newdimen\centeringwidth\centeringwidth=\graphwidth
  \advance\centeringwidth by\betweengraphswidth\divide\centeringwidth by 2
\newdimen\beforegraphamount\beforegraphamount=20pt
\newdimen\aftergraphamount\aftergraphamount=0pt
\def\skipgraphwidth{\hskip\graphwidth}
\def\skipbetweengraphs{\hskip\betweengraphswidth}
\def\skipcenteringwidth{\hskip\centeringwidth}
\def\skipbeforegraph{\vskip\beforegraphamount}
\def\skipaftergraph{\vskip\aftergraphamount}

\settabs\+
&{\skipgraphwidth}
&{\skipbetweengraphs}
&{\skipgraphwidth}
&{\skipbetweengraphs}
&{\skipgraphwidth}
&\cr

\+&{\epsfxsize=\graphwidth\epsfbox{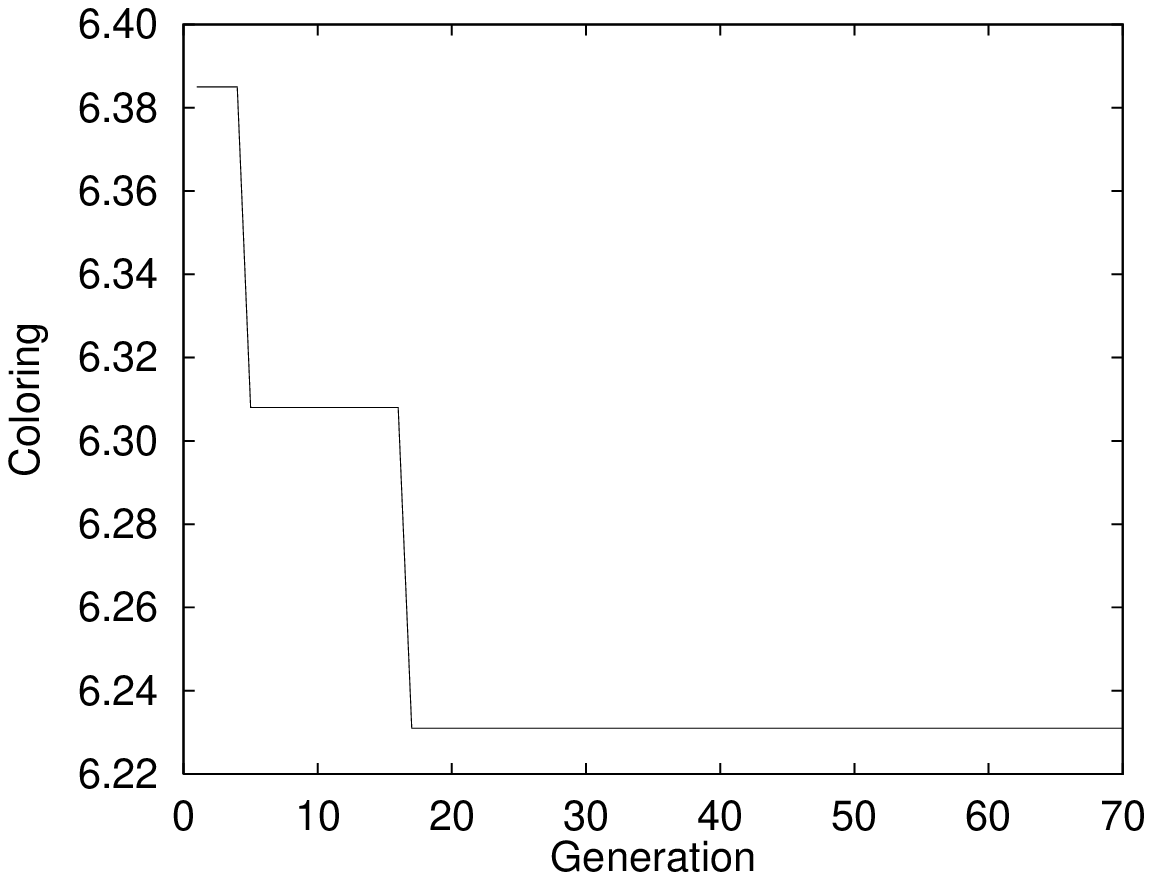}}
&&{\epsfxsize=\graphwidth\epsfbox{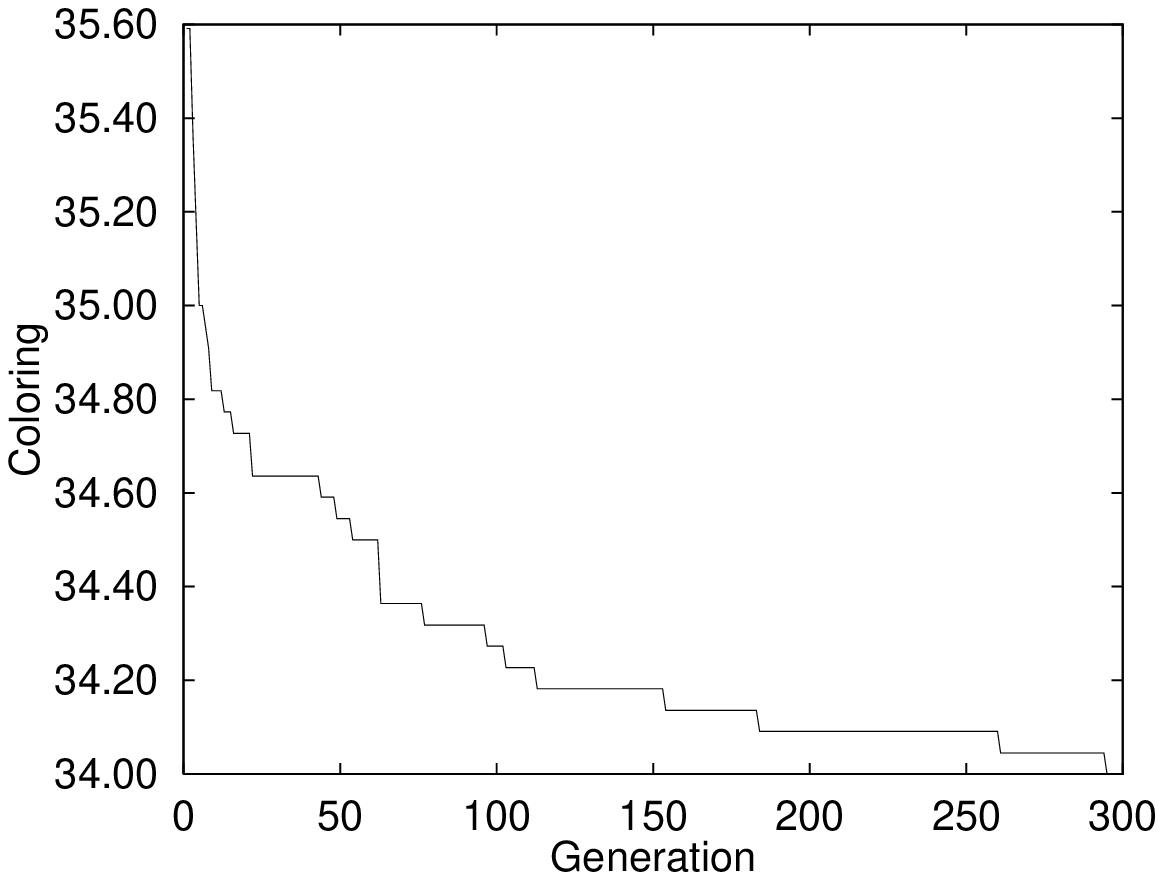}}
&&{\epsfxsize=\graphwidth\epsfbox{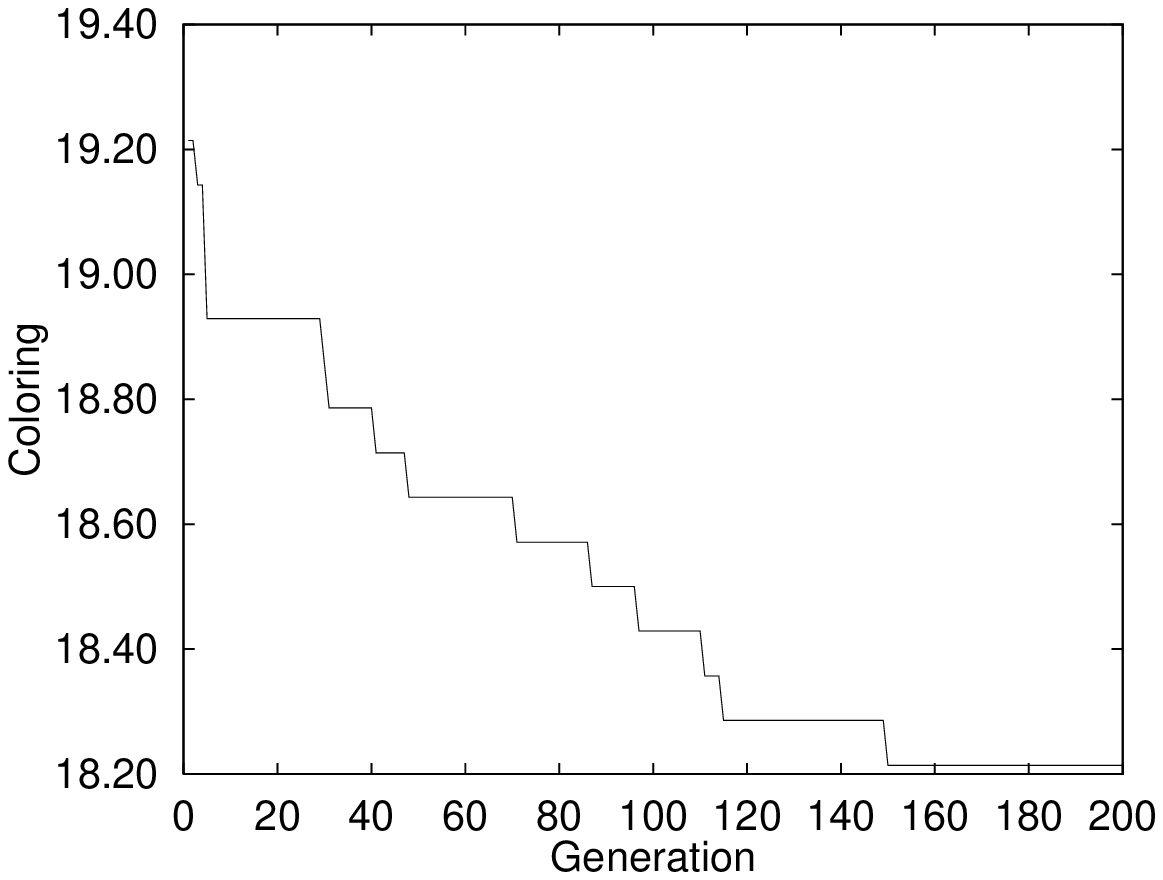}}
&\cr

\skipaftergraph
\+&\hfill{${\cal G}(125,0.02,0.09)$}\hfill
&&\hfill{${\cal G}(125,0.40,0.99)$}\hfill
&&\hfill{${\cal G}(197,0.10,0.39)$}\hfill
&\cr

\skipbeforegraph
\+&{\epsfxsize=\graphwidth\epsfbox{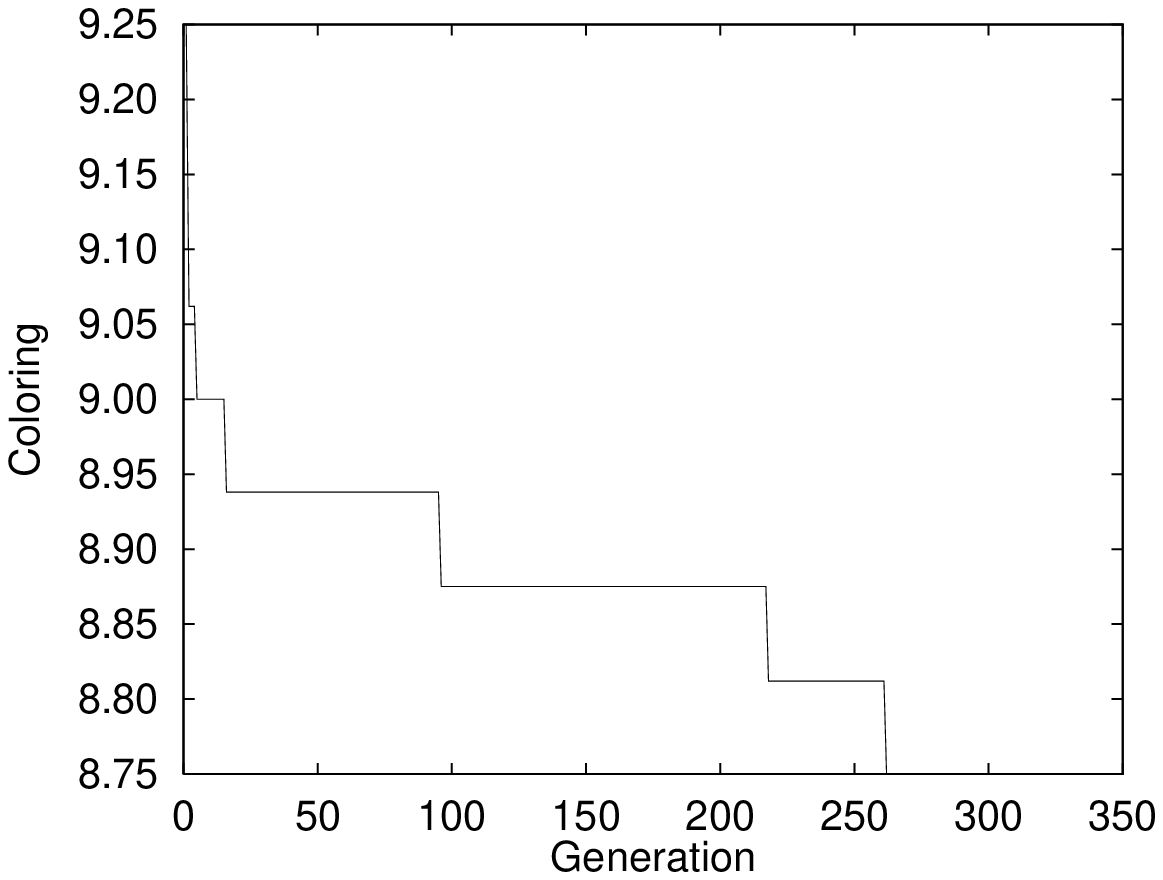}}
&&{\epsfxsize=\graphwidth\epsfbox{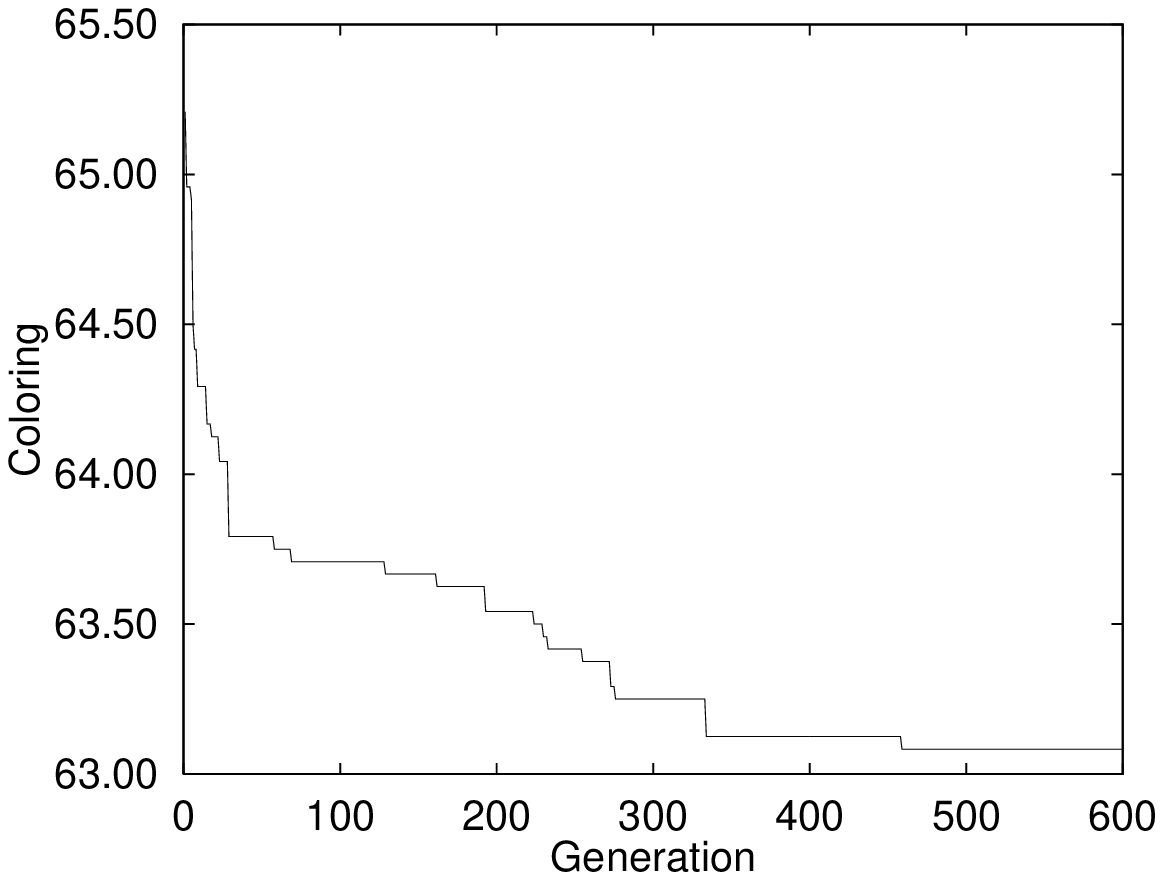}}
&&{\epsfxsize=\graphwidth\epsfbox{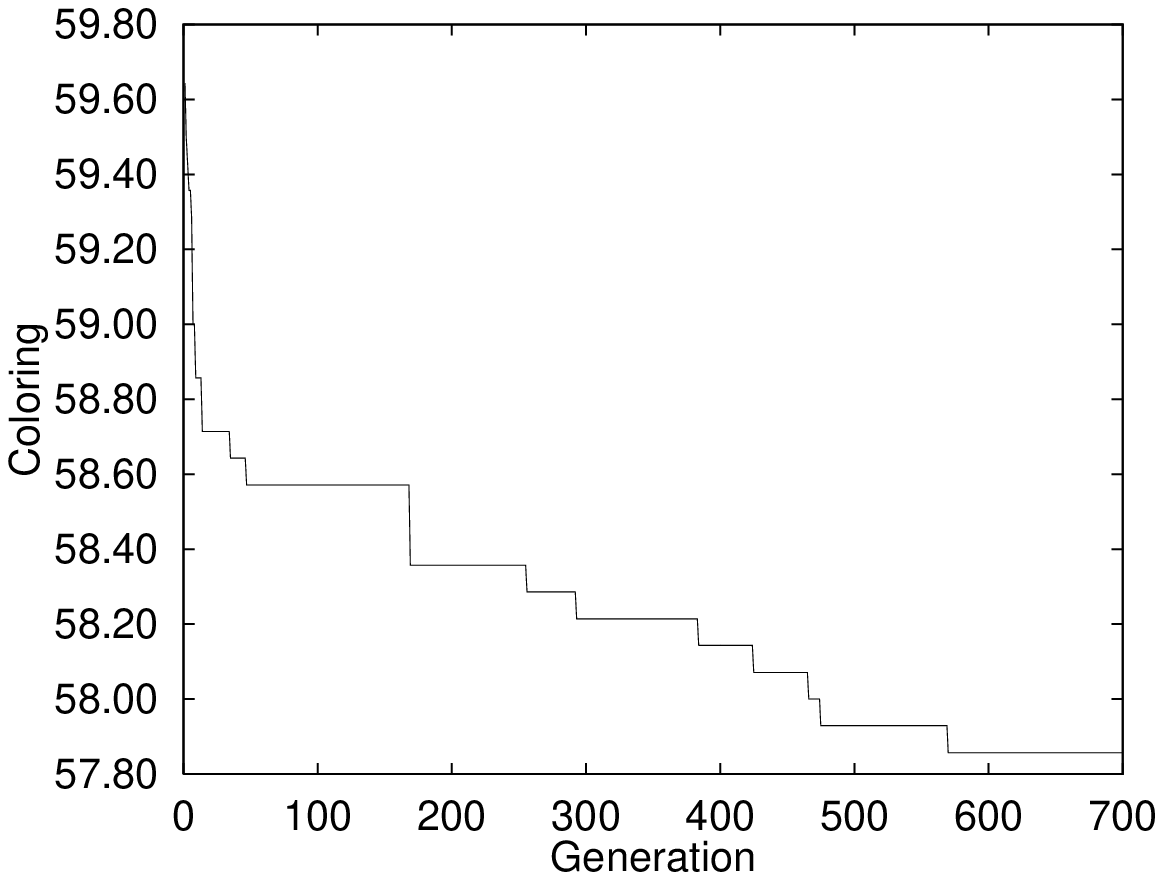}}
&\cr

\skipaftergraph
\+&\hfill{${\cal G}(250,0.02,0.09)$}\hfill
&&\hfill{${\cal G}(250,0.40,0.99)$}\hfill
&&\hfill{${\cal G}(300,0.40,0.99)$}\hfill
&\cr

\skipbeforegraph
\+&{\epsfxsize=\graphwidth\epsfbox{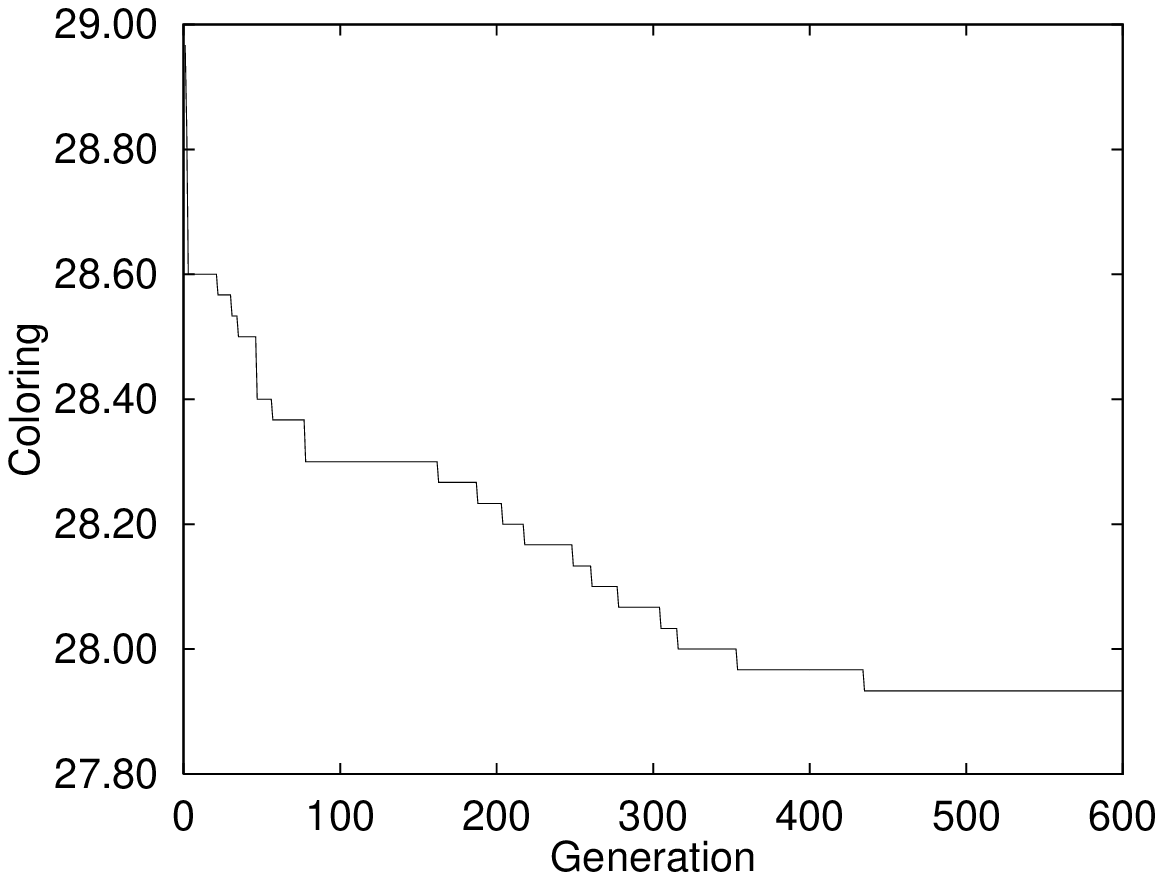}}
&&{\epsfxsize=\graphwidth\epsfbox{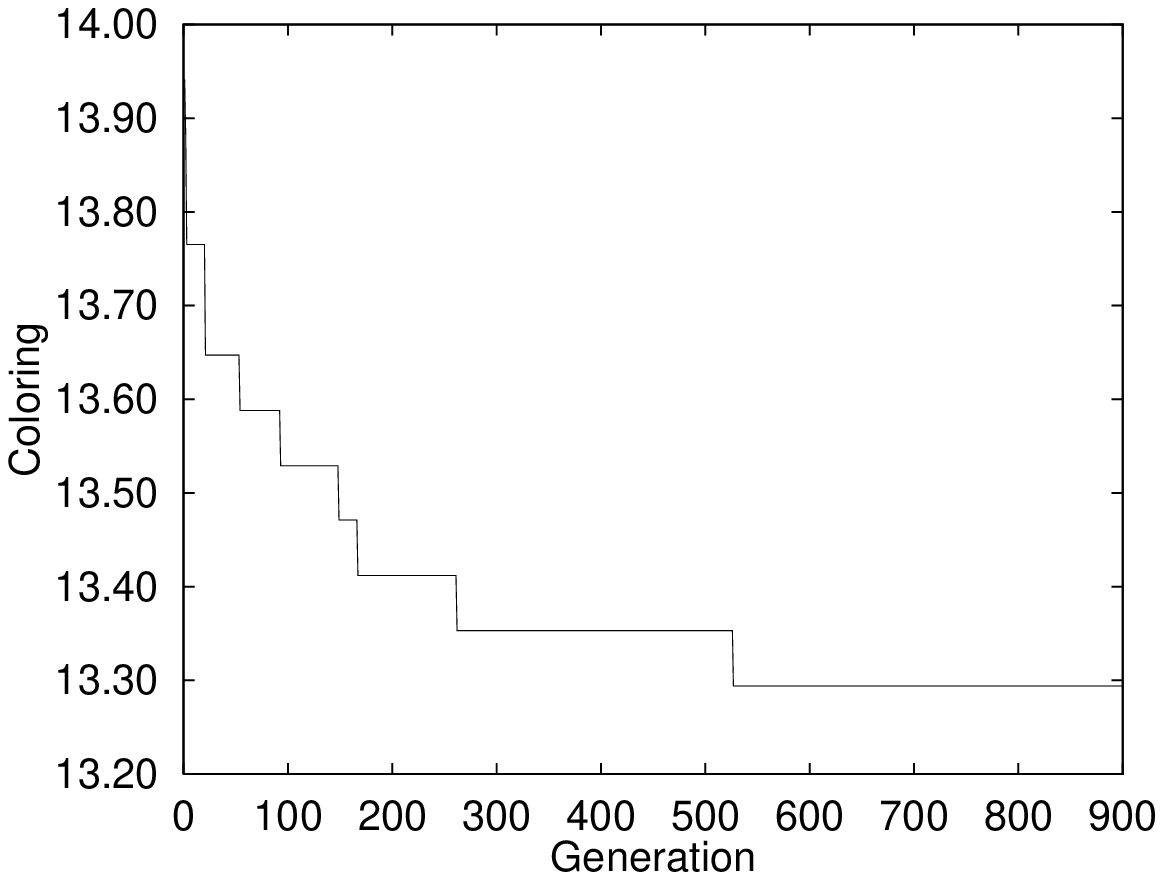}}
&&{\epsfxsize=\graphwidth\epsfbox{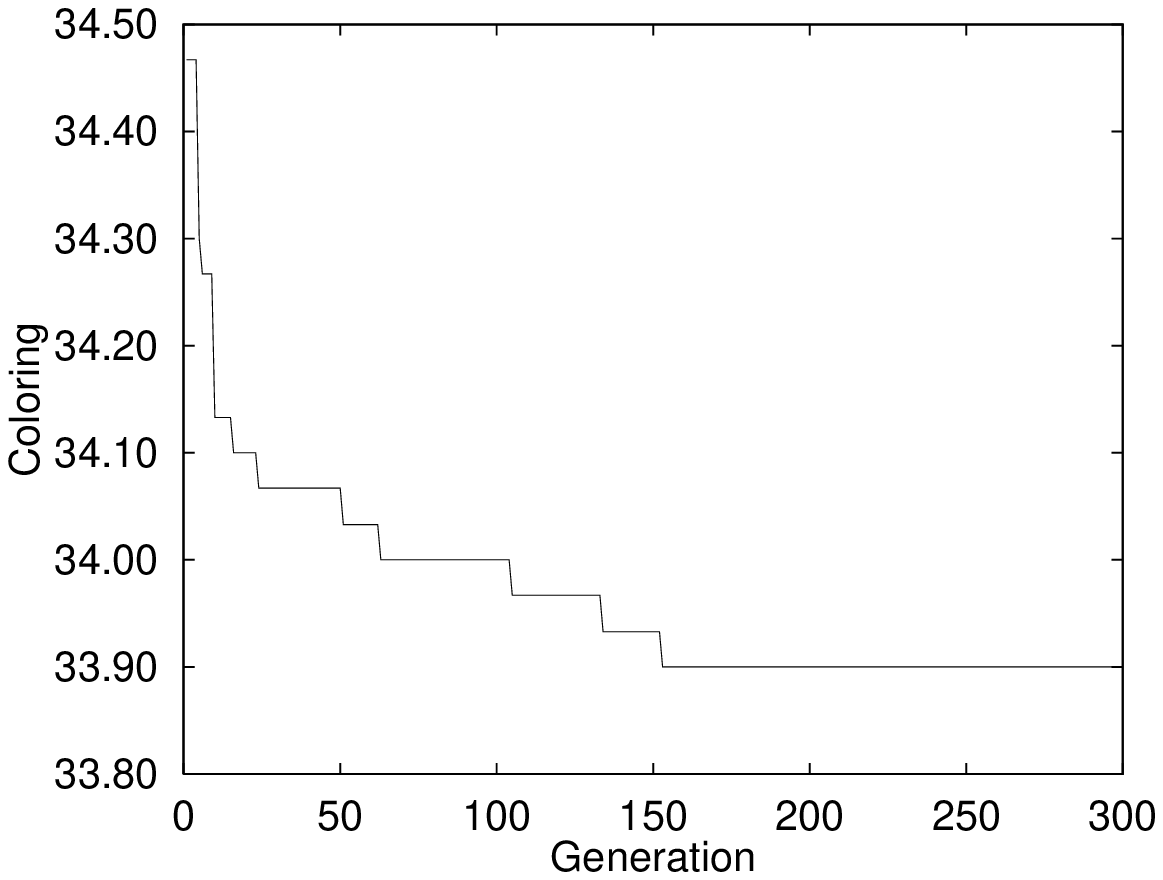}}
&\cr

\skipaftergraph
\+&\hfill{${\cal G}(352,0.10,0.39)$}\hfill
&&\hfill{${\cal G}(450,0.02,0.09)$}\hfill
&&\hfill{${\cal G}(450,0.10,0.39)$}\hfill
&\cr

\settabs\+
&{\skipcenteringwidth}
&{\skipgraphwidth}
&{\skipbetweengraphs}
&{\skipgraphwidth}
&\cr

\skipbeforegraph
\+&&{\epsfxsize=\graphwidth\epsfbox{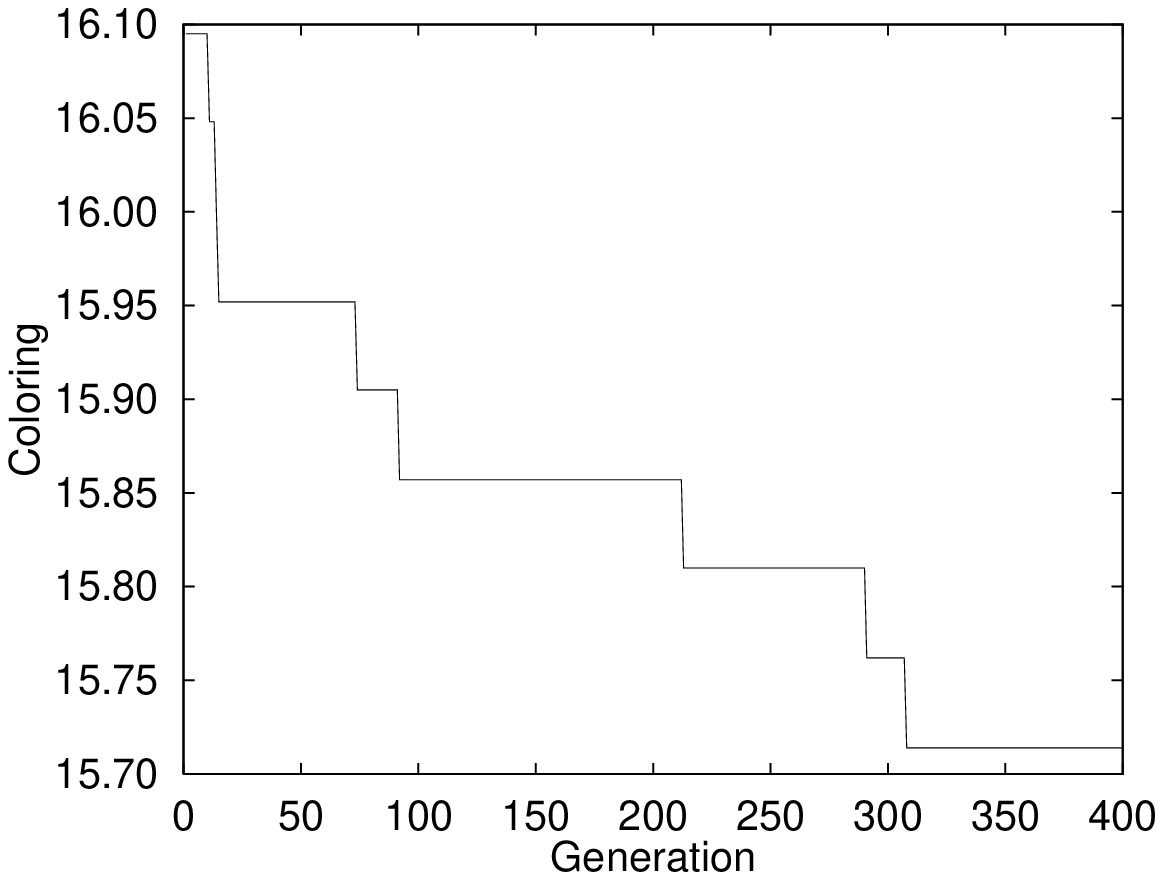}}
&&{\epsfxsize=\graphwidth\epsfbox{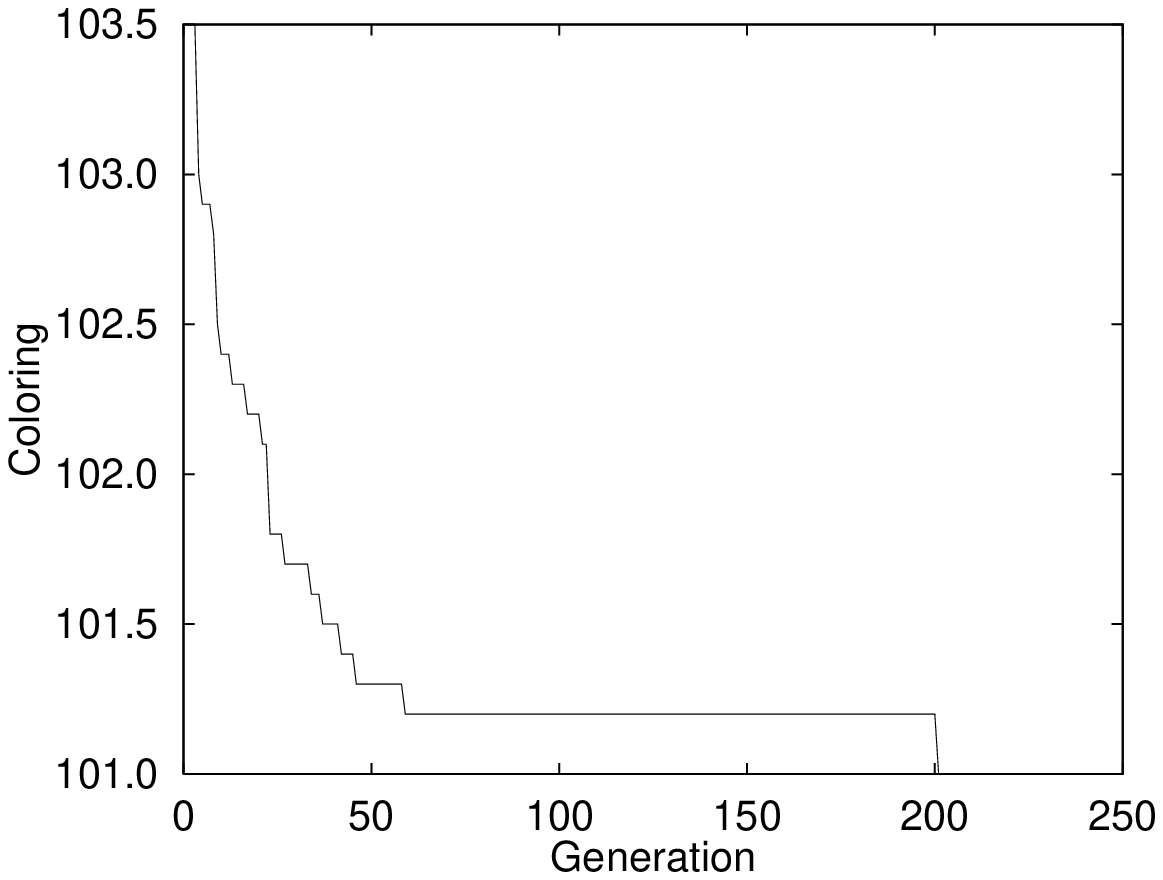}}
&\cr

\skipaftergraph
\+&&\hfill{${\cal G}(500,0.02,0.09)$}\hfill
&&\hfill{${\cal G}(500,0.40,0.99)$}\hfill
&\cr

}

}
$$
\bigskip
\centerline{{\bf Figure 3.} Convergence of {\sc Evolve\_P} on the target classes
of Table 1}
\endinsert

\topinsert
\centerline{{\bf Table 2.} Parameters for the experiments with {\sc Evolve\_P}}
\abovedisplayskip=0pt
$$
\vbox{
\settabs\+
&\skipbefore
&{${\cal G}(000,0.00,0.00)$}
&\skipbetween
&{$000$}
&\skipbetween
&{$0.00$}
&\skipbetween
&{$0.00$}
&\skipbetween
&{$0.00$}
&\skipbetween
&{$0.00$}
&\skipbetween
&{$000$}
&\skipbetween
&{$000$}
&\skipafter
&\cr

\+&\hrulefill
&\hrulefill
&\hrulefill
&\hrulefill
&\hrulefill
&\hrulefill
&\hrulefill
&\hrulefill
&\hrulefill
&\hrulefill
&\hrulefill
&\hrulefill
&\hrulefill
&\hrulefill
&\hrulefill
&\hrulefill
&\hrulefill
&\cr

\+&&\hfill{Target class}\hfill
&&\hfill{$\vert{\cal T}\vert$}\hfill
&&\hfill{$p_c$}\hfill
&&\hfill{$p_m$}\hfill
&&\hfill{$p_i$}\hfill
&&\hfill{$f$}\hfill
&&\hfill{$g$}\hfill
&&\hfill{$s$}\hfill
&&\cr

\vskip-\medskipamount
\+&\hrulefill
&\hrulefill
&\hrulefill
&\hrulefill
&\hrulefill
&\hrulefill
&\hrulefill
&\hrulefill
&\hrulefill
&\hrulefill
&\hrulefill
&\hrulefill
&\hrulefill
&\hrulefill
&\hrulefill
&\hrulefill
&\hrulefill
&\cr

\+&&{${\cal G}(125,0.02,0.09)$}
&&\skipd{$40$}
&&{$0.71$}
&&{$0.01$}
&&{$0.28$}
&&{$0.02$}
&&\hfill{$70$}
&&{$200$}
&&\cr

\+&&{${\cal G}(125,0.40,0.99)$}
&&\skipd{$60$}
&&{$0.71$}
&&{$0.01$}
&&{$0.28$}
&&{$0.02$}
&&{$300$}
&&{$100$}
&&\cr

\+&&{${\cal G}(197,0.10,0.39)$}
&&\skipd{$60$}
&&{$0.71$}
&&{$0.01$}
&&{$0.28$}
&&{$0.01$}
&&{$200$}
&&{$100$}
&&\cr

\+&&{${\cal G}(250,0.02,0.09)$}
&&\skipd{$60$}
&&{$0.71$}
&&{$0.01$}
&&{$0.28$}
&&{$0.01$}
&&{$350$}
&&{$100$}
&&\cr

\+&&{${\cal G}(250,0.40,0.99)$}
&&\skipd{$60$}
&&{$0.71$}
&&{$0.01$}
&&{$0.28$}
&&{$0.02$}
&&{$600$}
&&{$100$}
&&\cr

\+&&{${\cal G}(300,0.40,0.99)$}
&&\skipd{$60$}
&&{$0.71$}
&&{$0.01$}
&&{$0.28$}
&&{$0.01$}
&&{$700$}
&&{$100$}
&&\cr

\+&&{${\cal G}(352,0.10,0.39)$}
&&\skipd{$60$}
&&{$0.71$}
&&{$0.01$}
&&{$0.28$}
&&{$0.01$}
&&{$600$}
&&{$100$}
&&\cr

\+&&{${\cal G}(450,0.02,0.09)$}
&&{$100$}
&&{$0.71$}
&&{$0.01$}
&&{$0.28$}
&&{$0.02$}
&&{$900$}
&&{$100$}
&&\cr

\+&&{${\cal G}(450,0.10,0.39)$}
&&{$100$}
&&{$0.70$}
&&{$0.01$}
&&{$0.29$}
&&{$0.02$}
&&{$300$}
&&{$100$}
&&\cr

\+&&{${\cal G}(500,0.02,0.09)$}
&&\hfill{$40$}
&&{$0.70$}
&&{$0.01$}
&&{$0.29$}
&&{$0.02$}
&&{$400$}
&&{$100$}
&&\cr

\+&&{${\cal G}(500,0.40,0.99)$}
&&\hfill{$40$}
&&{$0.71$}
&&{$0.01$}
&&{$0.28$}
&&{$0.02$}
&&{$250$}
&&{$200$}
&&\cr

\vskip-\medskipamount
\+&\hrulefill
&\hrulefill
&\hrulefill
&\hrulefill
&\hrulefill
&\hrulefill
&\hrulefill
&\hrulefill
&\hrulefill
&\hrulefill
&\hrulefill
&\hrulefill
&\hrulefill
&\hrulefill
&\hrulefill
&\hrulefill
&\hrulefill
&\cr

}
$$
\endinsert

We show in Table 3 the best results obtained by {\sc Evolve\_AO} and
{\sc Evolve\_P} side by side with those obtained by the competing algorithms
we outlined earlier. For each of the graphs in Table 1, what Table 3 shows
is the value of its chromatic number (or an upper bound on it), when known by
design, and the colorings obtained by each of the algorithms to which the graph
was presented. For some algorithms this is given as an average over a certain
number of runs ($10$ for {\sc I\_Greedy}, $5$ for {\sc Par}, between $3$ and
$10$ for {\sc T\_gen\_1}, and $5$ for {\sc Evolve\_AO}), for some others as the
result of a single run ({\sc T\_B\&B}, {\sc E\_B\&B}, and {\sc Evolve\_P}). Yet
for other algorithms ({\sc Dist} and {\sc T\_Gen\_2}), what appears in Table 3
is the lowest value of $k$ for which a coloring by $k$ colors was obtained for
the corresponding graph, since those algorithms were tested by successively
decreasing the value of $k$ from some initial value. The entries in Table 3
corresponding to all algorithms but our own are either from [26] or, in the
case of {\sc T\_Gen\_2}, from [18].

While none of the heuristics appearing in Table 3 can be proclaimed strictly
best over all graphs, when we examine each graph individually many of the
heuristics can be said to do as well as the one that does best on that graph.
That is also the case with {\sc Evolve\_AO} and {\sc Evolve\_P}, which on
several graphs (those for which performance figures are given in bold typeface)
performed at least as well as the best performers on those same graphs,
occasionally even better. For the other graphs, often {\sc Evolve\_AO} and
{\sc Evolve\_P} miss the best performers very narrowly.

\topinsert
\centerline{{\bf Table 3.} Comparative performance on the benchmark graphs}
\abovedisplayskip=0pt
$$
\vbox{
\settabs\+
&\skipbefore
&{\sc flat300\_20\_0}
&\skipbetween
&{$\le 00$}
&\skipbetween
&{\sc I\_Greedy}
&\skipbetween
&{\sc T\_B\&B}
&\skipbetween
&{\sc Dist}
&\skipbetween
&{$00.0$}
&\skipbetween
&{\sc E\_B\&B}
&\skipafter
&\cr

\+&\hrulefill
&\hrulefill
&\hrulefill
&\hrulefill
&\hrulefill
&\hrulefill
&\hrulefill
&\hrulefill
&\hrulefill
&\hrulefill
&\hrulefill
&\hrulefill
&\hrulefill
&\hrulefill
&\hrulefill
&\cr

\+&&\hfill{$G$}\hfill
&&{$\chi(G)$}
&&{\sc I\_Greedy}
&&{\sc T\_B\&B}
&&{\sc Dist}
&&{\sc Par}
&&{\sc E\_B\&B}
&&\cr

\vskip-\medskipamount
\+&\hrulefill
&\hrulefill
&\hrulefill
&\hrulefill
&\hrulefill
&\hrulefill
&\hrulefill
&\hrulefill
&\hrulefill
&\hrulefill
&\hrulefill
&\hrulefill
&\hrulefill
&\hrulefill
&\hrulefill
&\cr

\+&&{\tt R125.1}\hfill
&&\hfill{$$}
&&\hfill{\skipd$5.0$}\hfill
&&\hfill{\skipd$5$}\hfill
&&\hfill{\skipd$5$}\hfill
&&\hfill{\skipd$5.0$}\hfill
&&\hfill{\skipd$5$}\hfill
&&\cr

\+&&{\tt R125.5}\hfill
&&\hfill{$$}
&&\hfill{$36.9$}\hfill
&&\hfill{$36$}\hfill
&&\hfill{$36$}\hfill
&&\hfill{$37.0$}\hfill
&&\hfill{$36$}\hfill
&&\cr

\+&&{\tt DSJC125.5}\hfill
&&\hfill{$$}
&&\hfill{$18.9$}\hfill
&&\hfill{$20$}\hfill
&&\hfill{$17$}\hfill
&&\hfill{$17.0$}\hfill
&&\hfill{$$}\hfill
&&\cr

\+&&{\tt R125.1c}\hfill
&&\hfill{$$}
&&\hfill{$46.0$}\hfill
&&\hfill{$46$}\hfill
&&\hfill{$46$}\hfill
&&\hfill{$46.0$}\hfill
&&\hfill{$46$}\hfill
&&\cr

\+&&{\tt mulsol.i.1}\hfill
&&\hfill{$$}
&&\hfill{$49.0$}\hfill
&&\hfill{$49$}\hfill
&&\hfill{$49$}\hfill
&&\hfill{$49.0$}\hfill
&&\hfill{$49$}\hfill
&&\cr

\+&&{\tt R250.1}\hfill
&&\hfill{$$}
&&\hfill{\skipd$8.0$}\hfill
&&\hfill{\skipd$8$}\hfill
&&\hfill{\skipd$8$}\hfill
&&\hfill{\skipd$8.0$}\hfill
&&\hfill{\skipd$8$}\hfill
&&\cr

\+&&{\tt R250.5}\hfill
&&\hfill{$$}
&&\hfill{$68.4$}\hfill
&&\hfill{$66$}\hfill
&&\hfill{$65$}\hfill
&&\hfill{$66.0$}\hfill
&&\hfill{$65$}\hfill
&&\cr

\+&&{\tt DSJC250.5}\hfill
&&\hfill{$$}
&&\hfill{$32.8$}\hfill
&&\hfill{$35$}\hfill
&&\hfill{$28$}\hfill
&&\hfill{$29.2$}\hfill
&&\hfill{$$}\hfill
&&\cr

\+&&{\tt R250.1c}\hfill
&&\hfill{$$}
&&\hfill{$64.0$}\hfill
&&\hfill{$65$}\hfill
&&\hfill{$64$}\hfill
&&\hfill{$64.0$}\hfill
&&\hfill{$64$}\hfill
&&\cr

\+&&{\tt flat300\_20\_0}\hfill
&&\hfill{$\le 20$}
&&\hfill{$20.2$}\hfill
&&\hfill{$39$}\hfill
&&\hfill{$20$}\hfill
&&\hfill{$20.0$}\hfill
&&\hfill{$$}\hfill
&&\cr

\+&&{\tt flat300\_26\_0}\hfill
&&\hfill{$\le 26$}
&&\hfill{$37.1$}\hfill
&&\hfill{$41$}\hfill
&&\hfill{$26$}\hfill
&&\hfill{$32.4$}\hfill
&&\hfill{$$}\hfill
&&\cr

\+&&{\tt flat300\_28\_0}\hfill
&&\hfill{$\le 28$}
&&\hfill{$37.0$}\hfill
&&\hfill{$41$}\hfill
&&\hfill{$31$}\hfill
&&\hfill{$33.0$}\hfill
&&\hfill{$$}\hfill
&&\cr

\+&&{\tt school1-nsh}\hfill
&&\hfill{$\le 14$}
&&\hfill{$14.1$}\hfill
&&\hfill{$26$}\hfill
&&\hfill{$20$}\hfill
&&\hfill{$14.0$}\hfill
&&\hfill{$$}\hfill
&&\cr

\+&&{\tt le450\_15a}\hfill
&&\hfill{$15$}
&&\hfill{$17.9$}\hfill
&&\hfill{$16$}\hfill
&&\hfill{$15$}\hfill
&&\hfill{$15.0$}\hfill
&&\hfill{$$}\hfill
&&\cr

\+&&{\tt le450\_15b}\hfill
&&\hfill{$15$}
&&\hfill{$17.9$}\hfill
&&\hfill{$15$}\hfill
&&\hfill{$15$}\hfill
&&\hfill{$15.0$}\hfill
&&\hfill{$15$}\hfill
&&\cr

\+&&{\tt le450\_15c}\hfill
&&\hfill{$15$}
&&\hfill{$25.6$}\hfill
&&\hfill{$23$}\hfill
&&\hfill{$15$}\hfill
&&\hfill{$16.6$}\hfill
&&\hfill{$$}\hfill
&&\cr

\+&&{\tt le450\_15d}\hfill
&&\hfill{$15$}
&&\hfill{$25.8$}\hfill
&&\hfill{$23$}\hfill
&&\hfill{$15$}\hfill
&&\hfill{$16.8$}\hfill
&&\hfill{$$}\hfill
&&\cr

\+&&{\tt DSJR500.1}\hfill
&&\hfill{$$}
&&\hfill{$12.0$}\hfill
&&\hfill{$12$}\hfill
&&\hfill{$12$}\hfill
&&\hfill{$12.0$}\hfill
&&\hfill{$12$}\hfill
&&\cr

\+&&{\tt DSJC500.5}\hfill
&&\hfill{$$}
&&\hfill{$58.6$}\hfill
&&\hfill{$65$}\hfill
&&\hfill{$49$}\hfill
&&\hfill{$53.0$}\hfill
&&\hfill{$$}\hfill
&&\cr

\+&&{\tt DSJR500.1c}\hfill
&&\hfill{$$}
&&\hfill{$85.0$}\hfill
&&\hfill{$87$}\hfill
&&\hfill{$85$}\hfill
&&\hfill{$85.3$}\hfill
&&\hfill{$$}\hfill
&&\cr

\vskip-\medskipamount
\+&\hrulefill
&\hrulefill
&\hrulefill
&\hrulefill
&\hrulefill
&\hrulefill
&\hrulefill
&\hrulefill
&\hrulefill
&\hrulefill
&\hrulefill
&\hrulefill
&\hrulefill
&\hrulefill
&\hrulefill
&\cr

}
$$
\endinsert

\topinsert
\centerline{{\bf Table 3 (continued)}}
\abovedisplayskip=0pt
$$
\vbox{
\settabs\+
&\skipbefore
&{\sc flat300\_20\_0}
&\skipbetween
&{$\le 00$}
&\skipbetween
&{\sc T\_Gen\_1}
&\skipbetween
&{\sc T\_Gen\_2}
&\skipbetween
&{\sc Evolve\_AO}
&\skipbetween
&{\sc Evolve\_P}
&\skipafter
&\cr

\+&\hrulefill
&\hrulefill
&\hrulefill
&\hrulefill
&\hrulefill
&\hrulefill
&\hrulefill
&\hrulefill
&\hrulefill
&\hrulefill
&\hrulefill
&\hrulefill
&\hrulefill
&\cr

\+&&\hfill{$G$}\hfill
&&{$\chi(G)$}
&&{\sc T\_Gen\_1}
&&{\sc T\_Gen\_2}
&&{\sc Evolve\_AO}
&&{\sc Evolve\_P}
&&\cr

\vskip-\medskipamount
\+&\hrulefill
&\hrulefill
&\hrulefill
&\hrulefill
&\hrulefill
&\hrulefill
&\hrulefill
&\hrulefill
&\hrulefill
&\hrulefill
&\hrulefill
&\hrulefill
&\hrulefill
&\cr

\+&&{\tt R125.1}\hfill
&&\hfill{$$}
&&\hfill{\skipd$5.0$}\hfill
&&\hfill{\skipd$$}\hfill
&&\hfill{\skipd$\bf 5.0$}\hfill
&&\hfill{\skipd$\bf 5$}\hfill
&&\cr

\+&&{\tt R125.5}\hfill
&&\hfill{$$}
&&\hfill{$35.6$}\hfill
&&\hfill{$$}\hfill
&&\hfill{$36.2$}\hfill
&&\hfill{$36$}\hfill
&&\cr

\+&&{\tt DSJC125.5}\hfill
&&\hfill{$$}
&&\hfill{$17.0$}\hfill
&&\hfill{$$}\hfill
&&\hfill{$17.2$}\hfill
&&\hfill{$20$}\hfill
&&\cr

\+&&{\tt R125.1c}\hfill
&&\hfill{$$}
&&\hfill{$46.0$}\hfill
&&\hfill{$$}\hfill
&&\hfill{$\bf 46.0$}\hfill
&&\hfill{$\bf 46$}\hfill
&&\cr

\+&&{\tt mulsol.i.1}\hfill
&&\hfill{$$}
&&\hfill{$49.0$}\hfill
&&\hfill{$$}\hfill
&&\hfill{$\bf 49.0$}\hfill
&&\hfill{$\bf 49$}\hfill
&&\cr

\+&&{\tt R250.1}\hfill
&&\hfill{$$}
&&\hfill{\skipd$8.0$}\hfill
&&\hfill{\skipd$$}\hfill
&&\hfill{\skipd$\bf 8.0$}\hfill
&&\hfill{\skipd$\bf 8$}\hfill
&&\cr

\+&&{\tt R250.5}\hfill
&&\hfill{$$}
&&\hfill{$69.0$}\hfill
&&\hfill{$$}\hfill
&&\hfill{$65.2$}\hfill
&&\hfill{$\bf 65$}\hfill
&&\cr

\+&&{\tt DSJC250.5}\hfill
&&\hfill{$$}
&&\hfill{$29.0$}\hfill
&&\hfill{$28$}\hfill
&&\hfill{$29.1$}\hfill
&&\hfill{$29$}\hfill
&&\cr

\+&&{\tt R250.1c}\hfill
&&\hfill{$$}
&&\hfill{$64.0$}\hfill
&&\hfill{$$}\hfill
&&\hfill{$\bf 64.0$}\hfill
&&\hfill{$\bf 64$}\hfill
&&\cr

\+&&{\tt flat300\_20\_0}\hfill
&&\hfill{$\le 20$}
&&\hfill{$20.0$}\hfill
&&\hfill{$$}\hfill
&&\hfill{$26.0$}\hfill
&&\hfill{$23$}\hfill
&&\cr

\+&&{\tt flat300\_26\_0}\hfill
&&\hfill{$\le 26$}
&&\hfill{$26.0$}\hfill
&&\hfill{$$}\hfill
&&\hfill{$31.0$}\hfill
&&\hfill{$28$}\hfill
&&\cr

\+&&{\tt flat300\_28\_0}\hfill
&&\hfill{$\le 28$}
&&\hfill{$33.0$}\hfill
&&\hfill{$31$}\hfill
&&\hfill{$33.0$}\hfill
&&\hfill{$\bf 29$}\hfill
&&\cr

\+&&{\tt school1-nsh}\hfill
&&\hfill{$\le 14$}
&&\hfill{$14.0$}\hfill
&&\hfill{$$}\hfill
&&\hfill{$\bf 14.0$}\hfill
&&\hfill{$20$}\hfill
&&\cr

\+&&{\tt le450\_15a}\hfill
&&\hfill{$15$}
&&\hfill{$15.0$}\hfill
&&\hfill{$$}\hfill
&&\hfill{$\bf 15.0$}\hfill
&&\hfill{$17$}\hfill
&&\cr

\+&&{\tt le450\_15b}\hfill
&&\hfill{$15$}
&&\hfill{$15.0$}\hfill
&&\hfill{$$}\hfill
&&\hfill{$\bf 15.0$}\hfill
&&\hfill{$17$}\hfill
&&\cr

\+&&{\tt le450\_15c}\hfill
&&\hfill{$15$}
&&\hfill{$16.0$}\hfill
&&\hfill{$15$}\hfill
&&\hfill{$16.0$}\hfill
&&\hfill{$25$}\hfill
&&\cr

\+&&{\tt le450\_15d}\hfill
&&\hfill{$15$}
&&\hfill{$16.0$}\hfill
&&\hfill{$$}\hfill
&&\hfill{$19.0$}\hfill
&&\hfill{$25$}\hfill
&&\cr

\+&&{\tt DSJR500.1}\hfill
&&\hfill{$$}
&&\hfill{$12.0$}\hfill
&&\hfill{$$}\hfill
&&\hfill{$\bf 12.0$}\hfill
&&\hfill{$\bf 12$}\hfill
&&\cr

\+&&{\tt DSJC500.5}\hfill
&&\hfill{$$}
&&\hfill{$51.0$}\hfill
&&\hfill{$48$}\hfill
&&\hfill{$52.5$}\hfill
&&\hfill{$59$}\hfill
&&\cr

\+&&{\tt DSJR500.1c}\hfill
&&\hfill{$$}
&&\hfill{$85.3$}\hfill
&&\hfill{$$}\hfill
&&\hfill{$\bf 85.0$}\hfill
&&\hfill{$\bf 85$}\hfill
&&\cr

\vskip-\medskipamount
\+&\hrulefill
&\hrulefill
&\hrulefill
&\hrulefill
&\hrulefill
&\hrulefill
&\hrulefill
&\hrulefill
&\hrulefill
&\hrulefill
&\hrulefill
&\hrulefill
&\hrulefill
&\cr

}
$$
\endinsert

\bigbeginsection 7. Concluding remarks

We have in this paper introduced two novel evolutionary formulations of the
graph coloring problem. The first formulation views the problem of finding a
graph's chromatic number as the problem of finding an acyclic orientation of
the graph according to which the longest directed path is shortest among all
acyclic orientations. Viewing the problem from this perspective immediately
provides the essential foundation for the evolutionary formulation, since each
acyclic orientation can be viewed as an individual whose fitness can be said to
be higher as its longest directed path is shorter. Despite this initial
simplicity, the design of appropriate evolutionary operators has relied on
sophisticated graph-theoretic arguments.

Our second formulation takes an entirely different route, and seeks to find a
program that will color any graph having a pre-specified number of nodes and
density within a pre-specified interval. This program is a permutation of
indices into a sequence of nodes that is previously agreed upon. The sequence we
have used contains nodes in nonincreasing order of degrees, having been loosely
inspired by the DSatur heuristic. But being only a reference sequence, any other
sequence could have been used as well. A program is then something like ``first
color the node whose degree is the third highest with the lowest available
color, then the node whose degree is the seventh highest, etc.,'' for example.
In this formulation, each individual is a program, its fitness being higher as
the average number of colors it requires to color all the graphs in a randomly
selected subset of graphs having that number of nodes and density in that
interval is smaller.

We have presented experimental results on some of the DIMACS benchmark graphs
and compared our algorithms' performances with those of the other heuristics
that we know to have been tested on those graphs as well. Our results indicate
generally competitive performance, sometimes reaching the best results obtained
by the other heuristics, occasionally better.

Notwithstanding these positive results, there is room, at least in principle,
for improvements to be attempted. For example, in more than one occasion we
have resorted to deterministic decisions to break symmetry, so it may be
worth checking whether introducing randomness at these points can have any
noticeable effect. Also, our two formulations are both purely evolutionary,
in the sense that at no point do they employ auxiliary heuristics, like for
example some form of local search. We think it may be worthwhile to investigate
such hybrid alternatives.

It is, however, in the context of {\sc Evolve\_P}, the algorithm that implements
our second formulation, that we believe the possibilities for extension and
improvement are the most challenging and interesting. For example, of all the
graphs that were presented to {\sc Evolve\_P} for coloring, the ones that
prompted the least satisfactory results were those whose nodes have degrees very
closely concentrated near the average, that is, graphs with very little variance
of node degree (this cannot be inferred from the data we have presented, but is
clear from a closer examination of the DIMACS files). We think this may be due
to the fact that perhaps this quantity was poorly represented by the graphs of
the training set. Because {\sc Evolve\_P} is strongly based on degrees and how
they relate to one another inside the graph, it is possible that targeting the
evolution not just at a certain number of nodes and a certain density interval,
but also at a certain interval of node-degree variance, might yield significant
improvements in some cases. This would, of course, call for a generator of
random graphs capable of controlling such variances in addition to densities,
which is to our knowledge also a topic for research.

\beginsection  Acknowledgments

The authors acknowledge partial support from CNPq, CAPES, the PRONEX initiative
of Brazil's MCT under contract 41.96.0857.00, and a FAPERJ BBP grant.

\bigbeginsection References

{\frenchspacing

\medskip
\item{1.} S. Arora and C. Lund,
``Hardness of approximations,''
in D. S. Hochbaum (Ed.),
{\it Approximation Algorithms for NP-Hard Problems},
PWS Publishing Company, Boston, MA, 1997, 399--446.

\medskip
\item{2.} G. Ausiello, P. Crescenzi, G. Gambosi, V. Kann,
A. Marchetti-Spaccamela, and M. Protasi,
{\it Complexity and Approximation},
Springer-Verlag, Berlin, Germany, 1999.

\medskip
\item{3.} W. Banzhaf, P. Nordin, R. E. Keller, and F. D. Francone,
{\it Genetic Programming: An Introduction},
Morgan Kaufmann Publishers, San Francisco, CA, 1998.

\medskip
\item{4.} V. C. Barbosa,
{\it An Atlas of Edge-Reversal Dynamics},
Chapman \& Hall/CRC, London, UK, 2000.

\medskip
\item{5.} V. C. Barbosa and E. Gafni,
``A distributed implementation of simulated annealing,''
{\it J. of Parallel and Distributed Computing\/} {\bf 6} (1989), 411--434.

\medskip
\item{6.} M. Bellare, O. Goldreich, and M. Sudan,
``Free bits, PCPs, and nonapproximability---towards tight results,''
{\it SIAM J. on Computing\/} {\bf 27} (1998), 804--915.

\medskip
\item{7.} J. A. Bondy and U. S. R. Murty,
{\it Graph Theory with Applications},
North-Holland, New York, NY, 1976.

\medskip
\item{8.} D. Br\'elaz,
``New methods to color vertices of a graph,''
{\it Comm. of the ACM\/} {\bf 22} (1979), 251--256.

\medskip
\item{9.} F. C. Chow and J. L. Hennessy,
``The priority-based coloring approach to register allocation,''
{\it ACM Trans. on Programming Languages and Systems\/} {\bf 12} (1990),
501--536.

\medskip
\item{10.} T. H. Cormen, C. E. Leiserson, R. L. Rivest, and C. Stein,
{\it Introduction to Algorithms, Second Edition},
The MIT Press, Cambridge, MA, 2001.

\medskip
\item{11.} J. C. Culberson and F. Luo,
``Exploring the $k$-colorable landscape with Iterated Greedy,''
in [26], 245--284.

\medskip
\item{12.} D. de Werra,
``An introduction to timetabling,''
{\it European J. of Operational Research\/} {\bf 19} (1985), 151--162.

\medskip
\item{13.} R. W. Deming,
``Acyclic orientations of a graph and chromatic and independence numbers,''
{\it J. of Combinatorial Theory B\/} {\bf 26} (1979), 101--110.

\medskip
\item{14.} A. E. Eiben, J. K. van der Hauw, and J. I. van Hemert,
``Graph coloring with adaptive evolutionary algorithms,''
{\it J. of Heuristics\/} {\bf 4} (1998), 25--46.

\medskip
\item{15.} C. Fleurent and J. Ferland,
``Genetic and hybrid algorithms for graph coloring,''
{\it Annals of Operations Research\/} {\bf 63} (1996), 437--461.

\medskip
\item{16.} C. Fleurent and J. A. Ferland,
``Object-oriented implementation of heuristic search methods for graph
coloring, maximum clique, and satisfiability,''
in [26], 619--652.

\medskip
\item{17.} D. B. Fogel,
{\it Evolutionary Computation},
IEEE Press, New York, NY, 1995.

\medskip
\item{18.} P. Galinier and J.-K. Hao,
``Hybrid evolutionary algorithms for graph coloring,''
{\it J. of Combinatorial Optimization\/} {\bf 3} (1999), 379--397.

\medskip
\item{19.} A. Gamst,
``Some lower bounds for a class of frequency assignment problems,''
{\it IEEE Trans. on Vehicular Technology\/} {\bf 35} (1986), 8--14.

\medskip
\item{20.} M. R. Garey and D. S. Johnson,
{\it Computers and Intractability: A Guide to the Theory of
NP-Completeness},
W. H. Freeman, New York, NY, 1979.

\medskip
\item{21.} M. R. Garey, D. S. Johnson, and H. C. So,
``An application of graph coloring to printed circuit testing,''
{\it IEEE Trans. on Circuits and Systems\/} {\bf CAS-23} (1976), 591--599.

\medskip
\item{22.} F. Glover and M. Laguna,
{\it Tabu Search},
Kluwer Academic Publishers, Boston, MA, 1997.

\medskip
\item{23.} F. Glover, M. Parker, and J. Ryan,
``Coloring by tabu branch and bound,''
in [26], 285--307.

\medskip
\item{24.} D. E. Goldberg,
{\it Genetic Algorithms in Search, Optimization and Machine Learning},
Addison-Wesley Publishing Company, Reading, MA, 1989.

\medskip
\item{25.} D. S. Johnson, C. R. Aragon, L. A. McGeoch, and C. Schevon,
``Optimization by simulated annealing: an experimental evaluation;
part ii, graph coloring and number partitioning,''
{\it Operations Research\/} {\bf 39} (1991), 378--406.

\medskip
\item{26.} D. S. Johnson and M. A. Trick (Eds.),
{\it Cliques, Coloring, and Satisfiability: Second DIMACS Implementation
Challenge},
American Mathematical Society, Providence, RI, 1996.

\medskip
\item{27.} R. M. Karp,
``Reducibility among combinatorial problems,''
in R. E. Miller and J. W. Thatcher (Eds.),
{\it Complexity of Computer Computations},
Plenum Press, New York, NY, 1972, 85--103.

\medskip
\item{28.} J. R. Koza,
{\it Genetic Programming},
The MIT Press, Cambridge, MA, 1992.

\medskip
\item{29.} F. T. Leighton,
``A graph coloring algorithm for large scheduling problems,''
{\it J. of Research of the National Bureau of Standards\/} {\bf 84} (1979),
489--505.

\medskip
\item{30.} G. Lewandowski and A. Condon,
``Experiments with parallel graph coloring heuristics and applications of
graph coloring,''
in [26], 309--334.

\medskip
\item{31.} B. D. McKay,
``Isomorph-free exhaustive generation,''
{\it J. of Algorithms\/} {\bf 26} (1998), 306--324.

\medskip
\item{32.} Z. Michalewicz and D. B. Fogel,
{\it How to Solve It: Modern Heuristics},
Springer-Verlag, Berlin, Germany, 2000.

\medskip
\item{33.} M. Mitchell,
{\it An Introduction to Genetic Algorithms},
The MIT Press, Cambridge, MA, 1996.

\medskip
\item{34.} C. Morgenstern,
``Distributed coloration neighborhood search,''
in [26], 335--357.

\medskip
\item{35.} Y. Saad,
{\it Iterative Methods for Sparse Linear Systems},
PWS Publishing Company, Boston, MA, 1996.

\medskip
\item{36.} E. C. Sewell,
``An improved algorithm for exact graph coloring,''
in [26], 359--373.

\medskip
\item{37.} R. P. Stanley,
``Acyclic orientations of graphs,''
{\it Discrete Mathematics\/} {\bf 5} (1973), 171--178.

\medskip
\item{38.} M. A. Trick,
``Appendix: Second DIMACS Challenge test problems,''
in [26], 653--657.

}

\bye